%% file: arxiv.tex
\documentclass[sigconf]{aamas} 

\input{body/package}
\usepackage{balance} % for balancing columns on the final page

\makeatletter
\gdef\@copyrightpermission{
  \begin{minipage}{0.2\columnwidth}
   \href{https://creativecommons.org/licenses/by/4.0/}{\includegraphics[width=0.90\textwidth]{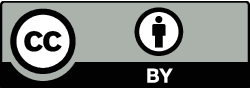}}
  \end{minipage}\hfill
  \begin{minipage}{0.8\columnwidth}
   \href{https://creativecommons.org/licenses/by/4.0/}{This work is licensed under a Creative Commons Attribution International 4.0 License.}
  \end{minipage}
  \vspace{5pt}
}
\makeatother

\setcopyright{ifaamas}
\acmConference[AAMAS '25]{Proc.\@ of the 24th International Conference
on Autonomous Agents and Multiagent Systems (AAMAS 2025)}{May 19 -- 23, 2025}
{Detroit, Michigan, USA}{Y.~Vorobeychik, S.~Das, A.~Nowé  (eds.)}
\copyrightyear{2025}
\acmYear{2025}
\acmDOI{}
\acmPrice{}
\acmISBN{}

\title[Bidirectional Distillation: A Mixed-Play Framework for Multi-Agent Generalizable Behaviors]{Bidirectional Distillation: A Mixed-Play Framework for Multi-Agent Generalizable Behaviors}

\author{Lang Feng}
\authornote{Equal contribution.}
\affiliation{
  \institution{Zhejiang University}
  \city{Hangzhou}
  \country{China}}
\email{langfeng@zju.edu.cn}

\author{Jiahao Lin}
\authornotemark[1]
\affiliation{
  \institution{Zhejiang University}
  \city{Hangzhou}
  \country{China}
  }
\email{22221159@zju.edu.cn}

\author{Dong Xing}
\affiliation{
  \institution{Zhejiang University}
  \city{Hangzhou}
  \country{China}
  }
\email{dongxing@zju.edu.cn}

\author{Li Zhang}
\affiliation{
  \institution{Zhejiang University}
  \city{Hangzhou}
  \country{China}}
\email{zhangli85@zju.edu.cn}

\author{De Ma}
\affiliation{
  \institution{Zhejiang University}
  \city{Hangzhou}
  \country{China}}
\email{made@zju.edu.cn}

\author{Gang Pan}
\authornote{Corresponding author.}
\affiliation{
  \institution{Zhejiang University}
  \city{Hangzhou}
  \country{China}}
\email{gpan@zju.edu.cn}

\begin{abstract}
\input{body/abstract}
\end{abstract}


\newcommand{\BibTeX}{\rm B\kern-.05em{\sc i\kern-.025em b}\kern-.08em\TeX}

\begin{document}

%%% The following commands remove the headers in your paper. For final 
%%% papers, these will be inserted during the pagination process.

\pagestyle{fancy}
\fancyhead{}

%%% The next command prints the information defined in the preamble.

\maketitle 

%%%%%%%%%%%%%%%%%%%%%%%%%%%%%%%%%%%%%%%%%%%%%%%%%%%%%%%%%%%%%%%%%%%%%%%%

\section{Introduction}
\input{body/introduction}
\section{Related Work}
\input{body/related_work}
\section{Setting and Background}
\input{body/background}
\section{Bidirectional Distillation}
\input{body/method}
\section{Experiment}
\input{body/experiment}
\section{Conclusion}
\input{body/conclusion}

\bibliographystyle{ACM-Reference-Format} 
\bibliography{sample}
\appendix
\input{body/appendix}
%%%%%%%%%%%%%%%%%%%%%%%%%%%%%%%%%%%%%%%%%%%%%%%%%%%%%%%%%%%%%%%%%%%%%%%%

\end{document}

%% file: body/package.tex
\usepackage{mathtools}
\usepackage{amsthm}
\usepackage{bm}
\usepackage{multirow}
\usepackage{xcolor}
\usepackage{hyperref}
\usepackage{booktabs}
\usepackage{pifont}
\usepackage{enumitem}
\usepackage{graphicx}
\usepackage{balance} % for balancing columns on the final page
\usepackage{algorithm}       
\usepackage{algorithmic}     

\usepackage{amsmath,amsfonts}
\usepackage{array}
\usepackage[caption=false,font=normalsize,labelfont=sf,textfont=sf]{subfig}
\usepackage{textcomp}
\usepackage{stfloats}
\usepackage{url}
\usepackage{verbatim}
\usepackage{graphicx}

%% file: body/abstract.tex
Population-population generalization is a challenging problem in multi-agent reinforcement learning (MARL), particularly when agents encounter unseen co-players. However, existing self-play-based methods are constrained by the limitation of inside-space generalization. In this study, we propose Bidirectional Distillation (BiDist), a novel mixed-play framework, to overcome this limitation in MARL. BiDist leverages knowledge distillation in two alternating directions: forward distillation, which emulates the historical policies' space and creates an implicit self-play, and reverse distillation, which systematically drives agents towards novel distributions outside the known policy space in a non-self-play manner. In addition, BiDist operates as a concise and efficient solution without the need for the complex and costly storage of past policies. We provide both theoretical analysis and empirical evidence to support BiDist’s effectiveness. Our results highlight its remarkable generalization ability across a variety of cooperative, competitive, and social dilemma tasks, and reveal that BiDist significantly diversifies the policy distribution space. We also present comprehensive ablation studies to reinforce BiDist’s effectiveness and key success factors. Source codes are available in the supplementary material.

%% file: body/introduction.tex
In recent years, significant strides~\cite{berner2019dota,jaderberg2019human,vinyals2019grandmaster,baker2019emergent} have been made in multi-agent reinforcement learning (MARL), with research exploring various avenues such as centralized training and decentralized execution (CTDE)~\cite{lowe2017maddpg,rashid2018qmix,yu2021mappo}, credit assignment~\cite{foerster2018counterfactual,wang2020qplex,kuba2022happo}, opponent modelling~\cite{foerster2018learning,wen2018probabilistic,dai2020r2} and more. Despite these advancements, MARL still lags behind supervised learning when it comes to the critical topic of generalization~\cite{hupkes2020compositionality}. The remarkable progress in supervised learning can be attributed, in part, to well-established protocols and sets of benchmarks that effectively assess algorithm generalization beyond the training data~\cite{deng2009imagenet,lecun2015deep}. In contrast, MARL algorithms are often developed to target a simplistic objective: excelling in the test environments that are distinguished from the training environments solely through random seeds~\cite{zhang2018dissection}. As a result, while state-of-the-art MARL algorithms can produce near-optimal agents for such homogeneous test environments, they are fragile and lack the ability to generalize when interacting with unseen scenarios.

\begin{figure}[t]
    \centering
    \includegraphics[width=0.82\columnwidth]{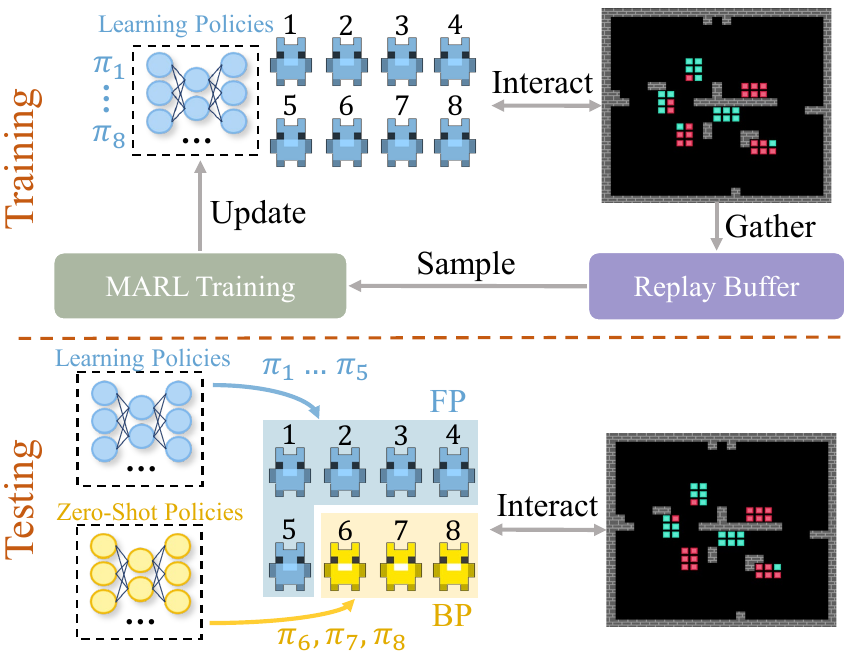}
    \caption{The training and testing phases of zero-shot co-player generalization task in MARL. FP and BP denote the focal population (blue) and background population (yellow) respectively.}
    \label{fig:example}
    \vspace{-0.1in}
\end{figure}
The generalization of MARL agents poses diverse challenges, encompassing variations in visual patterns~\cite{cobbe2019quantifying,gamrian2019transfer},  diverse dynamics~\cite{packer2018assessing}, zero-shot co-players~\cite{leibo2021scalable,agapiou2022melting} and etc. In our work, we concentrate on the generalization of a group of MARL agents when paired with a group of zero-shot co-players, framing it as a population-population generalization challenge. To elaborate, as depicted in Figure 1, all agents are trained jointly in the training phase, but during the testing phase, a random portion of agents is replaced with unseen agents called \emph{background}\footnote{we inherit the terminology ``background'' in~\cite{leibo2021scalable}.} population. The ultimate goal of this task is to ensure that the remaining training agents, referred to \emph{focal}\footnote{we inherit the terminology ``focal'' in~\cite{leibo2021scalable}.} population, continue to perform well with the zero-shot background population. Given that the behavior of a group of agents constitutes the environment of other agents, it is an effective means to assess the generalization ability of MARL agents when confronted with novel scenarios~\cite{leibo2021scalable}.

\begin{figure}[t]
    \centering
    \includegraphics[width=0.85\columnwidth]{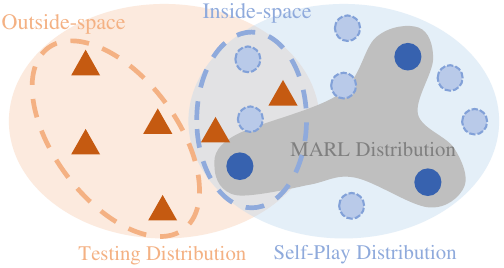}
    \caption{Inside-space and outside-space generalization. \protect{\raisebox{-.05cm}{\includegraphics[height=.3cm]{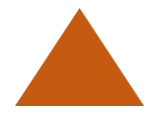}}} denotes the zero-shot policy, \protect{\raisebox{-.05cm}{\includegraphics[height=.3cm]{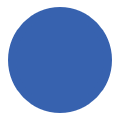}}} denotes the current training policy, and \protect{\raisebox{-.05cm}{\includegraphics[height=.3cm]{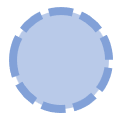}}} denotes the historical training policy.}
    \label{fig:inside-outside}
\end{figure}

A related but distinct area of research is ad-hoc teamwork~\cite{stone2010ad}, which aims to enable an individual agent to work effectively with unseen co-players. It has been a topic of extensive investigation~\cite{barrett2015cooperating,carroll2019utility,mirsky2020penny,gu2021online,rahman2021towards,charakorn2022generating,yuan2023survey,yan2024efficient}, but many of these methods are challenging to apply to complex population-population generalization tasks due to differences in training modalities and task types. 
Recent approaches~\cite{qiu2022rpm,jiang2024learning,nguyen2024diversifying} have explored the use of self-play mechanisms~\cite{heinrich2015fictitious,silver2018general} to address population-population generalization tasks. As illustrated in Figure~\ref{fig:inside-outside}, standard MARL only facilitates coordination among agents' \emph{current} policies during the training process, without accounting for other potential policies. Consequently, the interaction distribution of the agents remains highly limited (the grey region in Figure~\ref{fig:inside-outside}). In contrast, self-play broadens interaction diversity by leveraging historical policies, ideally being able to cover the entire distribution space of both \emph{current} and \emph{historical} training policies. This allows self-play to be particularly effective in addressing what we refer to as ``inside-space'' generalization (the blue region in Figure~\ref{fig:inside-outside}). 
However, the challenge arises when dealing with ``outside-space'' generalization where zero-shot policies include those never encountered during training. These distributions cannot be captured merely by reusing historical training policies, and in such cases, self-play-based methods fall short.

In this study, we propose a novel \emph{mixed-play framework}, which strategically deviates from the historical policy space to enrich the interaction diversity, particularly in the outside space. 
In this framework, a subset of agents is randomly detached from the training population, forming what we call the \emph{fictitious population}. This fictitious population acts as an imaginative background population for boosting the dynamics of agent-agent interactions during MARL training. 
At the core of this framework is a simple yet effective technique called \emph{Bidirectional Distillation (BiDist)}, based on knowledge distillation~\cite{hinton2015distilling}. The proposed BiDist comprises two alternating phases: forward distillation and reverse distillation, which aim to generate diverse distilled policies for the fictitious population. In the forward phase, the fictitious population’s policies are distilled from the historical policies, simulating implicit \emph{self-play} without the resource-intensive memorization of past policies observed in previous works~\cite{qiu2022rpm,jiang2024learning,nguyen2024diversifying}.
The reverse phase, by contrast, pushes the fictitious policies away from the historical policy space, encouraging novel behaviors outside of past patterns in a \emph{non-self-play} manner.
This capability allows for a systematic exploration of the outside-space, setting it apart from self-play. 

Our proposed BiDist has several appealing advantages: \textbf{(1)}~Our study justifies that BiDist’s reverse phase more effectively explores the outside-space in comparison to perturbation methods like entropy maximization. \textbf{(2)}~Our study leverages the concept of $\delta$-cover~\cite{sammut2017encyclopedia} and provide a theoretical justification to prove that $\delta$ is positively correlated with the generalization error bound. In our analysis, BiDist efficiently reduces $\delta$ as well as the generalization error bound. 
\textbf{(3)}~BiDist operates as a concise and efficient solution to the population-population generalization problem; there is no need for explicit storage of past policies. \textbf{(4)}~In our empirical evaluations, we test BiDist on various tasks including cooperation, competition, and social dilemma games. Our results substantiate the superiority of BiDist over baseline methods, and demonstrate its generalization and its ability to diversify the policy distribution space. Our work also presents comprehensive ablation studies to reinforce BiDist's robustness and success factors.

%% file: body/related_work.tex
Multi-agent reinforcement learning has been a subject of significant interest due to its challenging nature and practical applications. Various strong MARL algorithms, like value-based methods~\cite{rashid2018qmix,son2019qtran,wang2020qplex} and policy-based methods~\cite{lowe2017maddpg,iqbal2019actor,yu2021mappo,feng2023fp3o}, have achieved near-optimal performance. However, they provide no generalization guarantees when confronted with unseen co-players.

The issue of one agent working with unseen teammates without pre-coordination is the well-known ad-hoc teamwork~\cite{stone2010ad}. As a single-agent learning task, ad-hoc teamwork can be seen, in part, as a subtask of the population-population generalization challenge in MARL, where the focal population consists of just one agent. Although past research has explored various methods for ad-hoc teamwork~\cite{barrett2015cooperating,carroll2019utility,mirsky2020penny,gu2021online,rahman2021towards,charakorn2022generating,yuan2023survey,yan2024efficient,sarkar2024diverse}, they do not fully address the challenges of generalization in MARL, particularly in scenarios with multiple focal agents. Furthermore, these methods do not account for the arbitrary learning of other agents, making their applicability to population-population generalization tasks non-trivial.

Generalization in reinforcement learning has gained increasing attention in recent years. Several works on single-agent tasks have proposed to use data augmentation~\cite{raileanu2020automatic,kostrikov2020image}, regularization~\cite{farebrother2018generalization,cobbe2019quantifying} and randomization~\cite{lee2019network} to improve the robustness of agent. However, in the MARL domain, the generalization issue is more pluralistic~\cite{carion2019structured,hu2020other,strouse2021collaborating,mahajan2022generalization}. Here, our primary focus is on population-population generalization in MARL. \citet{vezhnevets2020options} proposed a hierarchical agent architecture that enables agents to generalize to unseen opponents, but it was specifically tailored for competitive games. \citet{zhao2023maximum} enhances pairwise diversity between agents and individual diversity through the derived Population Entropy bonus defined in the work. Other works such as \cite{lupu2021trajectory,yu2023learning,qiu2022rpm,jiang2024learning,nguyen2024diversifying} achieved diversity through self-play frameworks~\cite{tesauro1994td,heinrich2015fictitious,silver2018general}. For example, \citet{lupu2021trajectory} proposed Trajectory Diversity, which achieved population-based training but only for zero-shot coordination. \citet{jiang2024learning} incorporating multiple risk preferences diversified the strategy, forming a policy pool with different levels of performance and risk preferences. 
\citet{qiu2022rpm} proposed a ranked policy memory for population-population generalization task, addressing coordination, competition, and social dilemmas.
\citet{nguyen2024diversifying} proposed behavioral predictability as a criterion for selecting a diverse set of agents in the training pool.

However, these self-play-based methods are restricted to inside-space generalization and fail to capture outside-space distributions by relying solely on historical policy pools. Our method shares some similarities with the self-play but uses distilled policies to implicitly save the historical policies' space, eliminating the need for explicit policy pools. Furthermore, our mixed-play framework enables agents to explore and adapt to new possibilities beyond the confines of inside-space distribution, expanding their generalization capacity.

%% file: body/background.tex
In this part, we present the problem formulation of this work. Our primary focus is on the generalization of MARL agents in unseen scenarios where some co-players exhibit zero-shot behaviors~\cite{leibo2021scalable,agapiou2022melting}.

\paragraph{Problem Setting.}
We consider a decentralized partially observable Markov decision process (Dec-POMDP)~\cite{oliehoek2016concise}, which can be formally defined by a tuple $<\mathcal{N}, \mathcal{S}, \mathcal{A}, \mathcal{O}, P, \mathcal{R}, \gamma>$. Here, $\mathcal{N} \equiv \left\{ 1,\dots,N \right\} $ denotes the set of agents. $s \in \mathcal{S}$ is the global state of the environment. $\mathcal{A}=\mathcal{A}_i^N$ is the joint action space, where $\mathcal{A}_i$ denotes the action space of agent $i$. Similarly, $\mathcal{O}=\mathcal{O}_i^N$ is the joint local observation space, where $\mathcal{O}_i$ denotes the local observation spaces of agent $i$. The transition probability function $P:\mathcal{S}\times \mathcal{A} \times \mathcal{S} \rightarrow [0, 1]$ captures the dynamics of state transitions and $\mathcal{R}=r_i^N$ is the reward space with $r_i: \mathcal{S}\times \mathcal{A}\rightarrow \mathbb{R}$ denoting the individual reward function of agent $i$. $\gamma \in [0, 1)$ is the discount factor. Within each given state $s \in \mathcal{S}$, every agent $i$ has access solely to its own local observation $o_i\in \mathcal{O}_i$, and then makes a selection of an action $a_i\in \mathcal{A}_i$, drawn from its policy $\pi_i: \mathcal{O}_i\times\mathcal{A}_i\rightarrow [0, 1]$. We denote the joint action and joint policy as $\bm{a}=\{a_i\}_{i=1}^N$ and $\bm{\pi}=\{\pi_i\}_{i=1}^N$ respectively. Subsequently, each agent receives an individual reward $r_i(s, \bm{a})$, while the environment transitions to a new state based on the transition probability function. The policies of are parameterized by $\{\pi_{\theta_{i}}\}_{i=1}^N$ with parameter vectors $\bm{\theta}=\{\theta_{i}\}_{i=1}^N$.

\paragraph{Training protocol.}
The training environment is named as \emph{substrate} $\mathcal{Z}$~\cite{leibo2021scalable}, which encompasses all environmental factors beyond the $N$ agents themselves, involving the layout, physical rules, objects, and more. In the training phase, we let $N$ agents explore and train within this substrate, primarily following the aforementioned Dec-POMDP. Our goal is to train a $N$-player population $f$ with compatible policies for this substrate. The performance of the trained population on the substrate $\mathcal{Z}$ is measured by the expected per-capita return:
\begin{align}
    % \label{eq:per_capita_return_train}
    \bar{R}(f|\mathcal{Z})=\frac{1}{N}\sum_{i=1}^N \mathbb{E}_{\pi_1\sim f,\dots,\pi_N\sim f}\mathbb{E}\left[ \sum_{t=0}^{\infty} \gamma^t r_i(s_t, \bm{a}_t) \right], 
\end{align}
where $t$ is the time step. $\mathbb{E}_{\pi_1\sim f,\dots,\pi_N\sim f}$ denotes that policies are independently sampled from the trained population $f$. 
% $\mathbb{E}_{s,\bm{a}}$ is the abbreviated form of $\mathbb{E}_{s_{0:\infty},\bm{a}_{0:\infty}}$.

\paragraph{Testing protocol.}
The testing environment is referred to as the \emph{scenario} $\mathcal{Z}^{\prime}$~\cite{leibo2021scalable}, which consists of the substrate $\mathcal{Z}$ and the background population $g$. A binary vector $\bm{c}=\{c_1,\dots,c_N\}\in\{0,1\}^N$ with $m$ ones and $N-m$ zeros is utilized to decide whether agent $i$ belongs to the \emph{focal} population ($c_i=1$) or \emph{background} population ($c_i=0$). The background population $g$ is a set of $N-m$ unseen agents controlled by pre-trained policies, which constitutes an integral part of the environment. To measure the generalization performance of the trained focal population, the expected per-capita return is adopted as the evaluative criterion, with rewards of the background population excluded. Specifically, the expected focal per-capita return is defined as:
\begin{align}
    \bar{R}(f|\mathcal{Z}^{\prime})=&\bar{R}(f|\mathcal{Z}, \bm{c}, g) \nonumber\\
    =&\frac{1}{m}\sum_{i=1}^N c_i \mathbb{E}_{\pi_1\sim h_1,\dots,\pi_N\sim h_N}\mathbb{E}\left[ \sum_{t=0}^{\infty} \gamma^t r_i(s_t, \bm{a}_t) \right], 
\end{align}
where $h_i=f^{c_i}\cdot g^{1-c_i}$ that is used to distinguish the focal and background population.

%% file: body/method.tex
\begin{figure*}[t]
    \centering
    \includegraphics[width=1\textwidth]{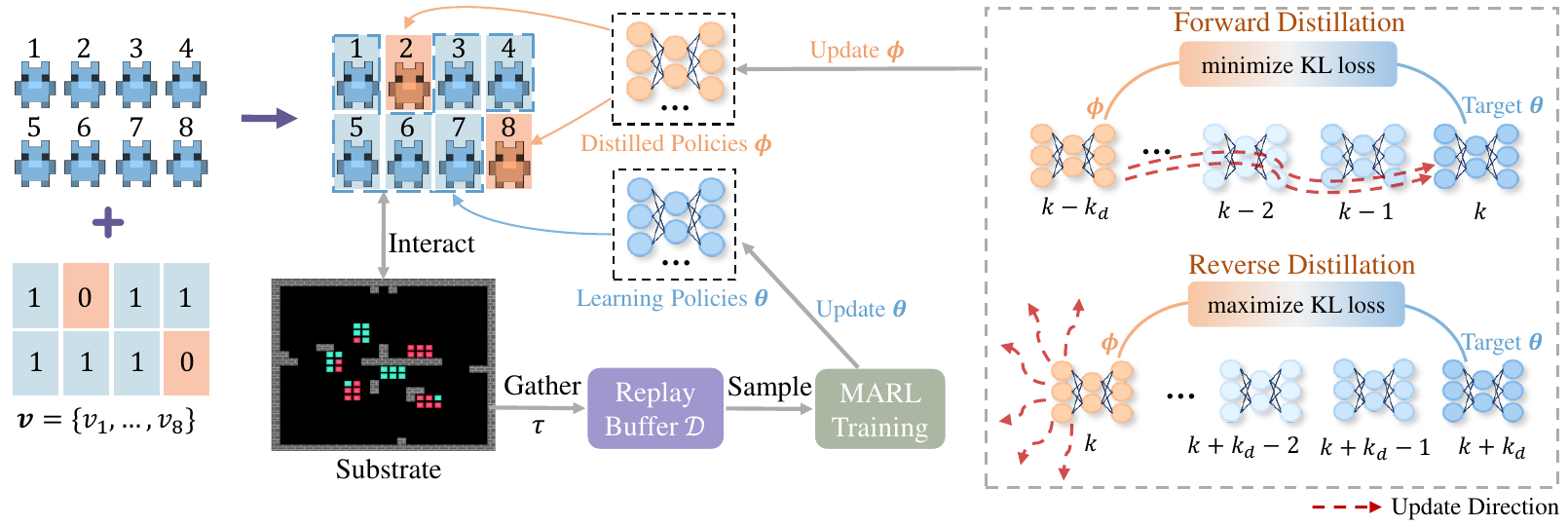}
    \caption{Illustration of BiDist in an 8-agent task. Based on vector $\bm{v}$, agents are divided into two populations: trained (blue) and fictitious (orange) populations. They respectively adopt the learning policies (blue) and the distilled policies (orange) to gather trajectories in the substrate. The distilled policies are updated by forward and reverse distillations alternately at an interval $k_d$. The diagram shows the iterations from $k-k_d$ to $k+k_d$, a complete alternating cycle, where forward distillation occurs at iteration $k$; reverse distillation occurs at iteration $k+k_d$; the remaining iterations do not involve any distillations.}
    \label{fig:bidirectional-distillation}
\end{figure*}
In response to the dual challenges posed by inside-space and outside-space generalization in population-population scenarios, we introduce \emph{Bidirectional Distillation (BiDist)} method. It functions as a sophisticated mixed-play framework and addresses these challenges through a two-pronged approach: forward distillation and reverse distillation. By alternating between these phases, BiDist enables agents to both retain essential self-play knowledge (addressing inside-space generalization) and push beyond historical boundaries to explore new distribution (addressing outside-space generalization).

In the following sections, we will \textbf{(1)} delve into the construction of fictitious population, \textbf{(2)} outline the process of forward distillation and reverse distillation, \textbf{(3)} provide a theoretical analysis of BiDist's effectiveness in reducing the generalization error bound, and \textbf{(4)} instantiate the whole workflow of BiDist.

\subsection{Random Fictitious Population}
% To enhance the robustness of agents when encountering unseen co-players, it is crucial to ensure that $N$-agents can produce diverse policy combinations and multi-agent interactions when trained in the substrate. 
We first introduce the \emph{fictitious} population $\tilde{g}$ to act as an \emph{imaginary background population} $g$ during the training phase.
It consists of a subset of agents randomly detached from the trained population $f$. Specifically, in order to generate $\tilde{g}$ in each training iteration, we use a binary vector $\bm{v}=\{v_1,\dots,v_N\}\in\{0,1\}^N$ to determines whether agent $i$ belongs to the fictitious population ($v_i=0$) or the trained population ($v_i=1)$. The vector $\bm{v}$ is sampled from the following probability distribution.
\begin{align}
    &v_i \sim \mathfrak{B}(\bm{v}_{<i}; p, N) \\
    &P(v_i) = p^{\delta_i}\cdot\mathbb{I}(v_i=1) + (1-p^{\delta_i})\cdot\mathbb{I}(v_i=0)\\
    &\delta_i = 1-\mathbb{I}(i=N)\cdot\mathbb{I}(\bm{v}_{<i}=\mathbf{0}_{<i}),
\end{align}
where $\mathbb{I}(\cdot)$ is the indicator function and $\mathbf{0}_{<i}=\{0\}_{\times{i-1}}$ denotes a vector of length $i-1$ with all elements set to 0. In most cases when $\delta_i=1$, $\mathfrak{B} (\bm{v}_{<i}; p, N)$ is a standard \emph{Bernoulli distribution}, indicating that each agent $i$ has a probability of $p$ to belong to the trained population $f$. However, there is an exception when agents $1,..., N-1$ all belong to the fictitious population $\tilde{g}$. In such a case, $\mathfrak{B} (\bm{v}_{<i}; p, N)$ degenerates into a \emph{deterministic} distribution, ensuring that the last agent $N$ definitely belongs to the trained population $f$. This safeguard prevents all agents from being assigned to the fictitious population $\tilde{g}$, which holds no meaning as there are no scenarios where all agents act as background agents.

The agents within the fictitious population $\tilde{g}$ do \emph{not} employ the learning policies $\{\pi_{\theta_i}\}_{i=1}^N$ during their interactions in the substrate; instead, they utilize the distilled policies $\{\pi_{\phi_i}\}_{i=1}^N$ with parameter vectors $\bm{\phi}=\{\phi _i\}_{i=1}^N$ as follows:
\begin{equation}
    \tau \sim \texttt{GatherTrajectories}(\mathcal{Z}, \underbrace{\{\pi_{\theta_i}\}_{v_i=1}}_{\text{Population } f}, \underbrace{\{\pi_{\phi_i}\}_{v_i=0}}_{\text{Population } \tilde{g}}).
\end{equation}
Here, $\tau=\{s,o_i,a_i,r_i\}_{i=1}^N$ denotes an experience pair, which is subsequently pushed into a replay buffer $\mathcal{D}$, as illustrated in Figure~\ref{fig:bidirectional-distillation}.
The distilled policies $\{\pi_{\phi_i}\}_{i=1}^N$ are generated by forward distillation and reverse distillation of BiDist, which we will introduce later. 
By doing so, fictitious population $\tilde{g}$ ensures that $N$-agents can produce diverse multi-agent policy combinations and distributions when trained in the substrate.

\subsection{Forward Distillation}
Indeed, the policies or behaviors of fictitious population $\tilde{g}$ are desired to exhibit distinctions to diversify the multi-agent interactions during gameplay in the substrate. As such, self-play methods like RPM~\cite{qiu2022rpm} explicitly store a number of agents' historical learning policies and allow agents to load these stored policies to cover as many decision-making patterns as possible. This approach is effective in addressing inside-space challenges, as discussed in Figure~\ref{fig:inside-outside}.

In this study, we propose a more concise solution by utilizing knowledge distillation~\cite{hinton2015distilling} to implicitly preserve the spaces of historical learning policies and avoid excessive resource consumption. Specifically, this entails employing the distilled policy networks $\{\pi_{\phi_i}\}_{i=1}^N$ to mimic the distributions of the learning policy networks $\{\pi_{\theta_i}\}_{i=1}^N$ during the whole training process. The training objective is to minimize the KL divergence between the distilled policy networks and the learning policy networks as follows:
\begin{align}
    \label{eq:forward_distillation}
    \mathcal{L}^{\rm KL}\left(\bm{\phi}\right)&=\frac{1}{N}\sum\limits_{i=1}^N\mathbb{E}_{o_i\sim \mathcal{D}}\left[D_{\rm KL}\left(\pi_{\theta_i}(\cdot|o_i) || \pi_{\phi_i}(\cdot|o_i)\right) \right]\nonumber \\
    \bm{\phi}&\leftarrow \bm{\phi}-\eta_f \cdot \nabla_{\bm{\phi}}\mathcal{L}^{\rm KL}\left(\bm{\phi}\right),
\end{align}
where $D_{\rm KL}$ denotes the KL divergence between two distributions. $\eta_f$ is the learning rate of forward distillation. Equation~(\ref{eq:forward_distillation}) ensures that the distilled policies closely emulate the learning policies and make similar decisions, thus termed as \emph{forward distillation}. It is worth noting that the learning policy networks are continuously updated at every single iteration, while forward distillation is periodically executed at a specified interval of $k_d$ iterations. The interval $k_d$ enables distilled policy networks to act as lagged representations of the continuously updated learning policy networks until the next distillation update turn, as shown in Figure~\ref{fig:bidirectional-distillation}.

\subsection{Reverse Distillation}
Nevertheless, forward distillation is inherently limited by the domain of historical policies, much like self-play methods. In certain scenarios, such as the chicken game, while the trained teammates might have learned to prefer the ``dove'' strategy, the unseen teammates might lean towards the ``hawk'' strategy. This highlights the importance of inducing \emph{preference shifts} for fictitious population $\tilde{g}$ to diversify the interactions during the training. Formally, given an observation, we hope $a^{\text{prefer}}_{\theta_i}\neq a^{\text{prefer}}_{\phi_i}$ is achievable, where $a^{\text{prefer}}_{\theta_i}=\arg\max_{a_i}\pi_{\theta_i}(a_i|o_i)$ and $a^{\text{prefer}}_{\phi_i}=\arg\max_{a_i}\pi_{\phi_i}(a_i|o_i)$ are the preferred actions of trained population and fictitious population, respectively.

Although there are various perturbation ways that can enhance the exploration, such as noise injection or entropy maximization, they might not lead to shifts in decision-making preferences. Entropy maximization, for instance, can only push the distilled policies to its upper bound $\mathcal{H}^{\max}(\pi_{\phi_i}(\cdot|o_i))$ to achieve the maximum uncertainty (i.e., $\forall a_i\sim \mathcal{A}_i$, $\pi_{\phi_i}(a_i|o_i)=\frac{1}{n}$), but it fails to enforce preference shifts. Here, we present the following proposition to show that KL maximization can drive further shifts in decision-making preferences compared to entropy maximization.
\begin{proposition}
    \label{proposition:preference_shift}
    Given the observation $o_i$ and a distilled policy $\pi_{\phi_i}$ that attains the upper bound $\mathcal{H}^{\max}(\pi_{\phi_i}(\cdot|o_i))$ of entropy maximization, there exists an update margin for KL divergence maximization that enables $\pi_{\phi_i}(a^{\text{prefer}}_{\theta_i}|o_i)$ to continue decreasing, resulting in the preference shift $a^{\text{prefer}}_{\theta_i}\neq a^{\text{prefer}}_{\phi_i}$.
\end{proposition}
The proof is provided in Appendix~A. 
% Appendix~\ref{appendix:proofs}. 
Building on this, we propose to use \emph{reverse distillation} as follows:
\begin{align}
    \label{eq:reverse_distillation}
    \bm{\phi}&\leftarrow \bm{\phi}+\eta_r \cdot \nabla_{\bm{\phi}}\mathcal{L}^{\rm KL}\left(\bm{\phi}\right),
\end{align}
where $\eta_r$ is the learning rate of reverse distillation. This process aims to push the distilled policies away from the historical policies' distributions by increasing the KL divergence. In this way, the distilled policies can capture action preferences that differ from the learning policies, enabling them to explore and generalize beyond the inside-space distribution. 

Throughout the entire training process, forward distillation and reverse distillation are executed alternately at regular intervals, every $k_d$ iteration, hence named Bidirectional Distillation. BiDist accommodates retention and innovation, ensuring that the distilled policies inherit the strengths of their predecessors (self-play policies) while being empowered to explore and adapt to new outside-space possibilities.

\subsection{Theoretical Analysis}
\input{body/theoretical}

\subsection{Policy Iteration}
In this part, we present the policy iteration of BiDist in Algorithm~\ref{alg:bidirectional_distillation}. Based on this, together with Figure~\ref{fig:bidirectional-distillation}, we outline the whole workflow of BiDist.
\begin{algorithm}[tb]
    \caption{Policy iteration of BiDist}
    \label{alg:bidirectional_distillation}
    \textbf{Input}: Substrate $\mathcal{Z}$, probability $p$, distillation interval $k$, and learning rates $\eta_f$ and $\eta_r$.\\
    \textbf{Initialize}: $\bm{\theta}$ of learning policy networks and $\bm{\phi}$ of distilled policy networks.
    \begin{algorithmic}[1] %[1] enables line numbers
    \FOR{$k=1,2,...$ until convergence}
    \STATE Randomize agents' indexes.
    \FOR{$i=1,...,N$}
    \STATE Sample $v_i$ from $\mathfrak{B} (\bm{v}_{<i}; p, N)$.
    \ENDFOR 
    \STATE Partition populations $f$ and $\tilde{g}$ with $\{v_1,...,v_N\}$.
    \STATE $\mathcal{D}\leftarrow\texttt{GatherTrajectories}(\mathcal{Z}, f, \tilde{g})$.
    \STATE Update the policies by $\bm{\theta}\leftarrow \arg\max_{\bm{\theta}}\mathcal{J}(\bm{\theta})$.
    \IF {$k$ is an odd multiple of $k_d$}
    \STATE Forward distillation: $\bm{\phi}\leftarrow \bm{\phi}-\eta_f \cdot \nabla_{\bm{\phi}}\mathcal{L}^{\rm KL}\left(\bm{\phi}\right)$
    \ELSIF {$k$ is an even multiple of $k_d$}
    \STATE Reverse distillation: $\bm{\phi}\leftarrow \bm{\phi}+\eta_r \cdot \nabla_{\bm{\phi}}\mathcal{L}^{\rm KL}\left(\bm{\phi}\right)$
    \ENDIF
    \ENDFOR
    \RETURN $\bm{\theta}$ and $\bm{\phi}$
    \end{algorithmic}
\end{algorithm}

\emph{Lines 2-6:} At the beginning of each policy iteration $k$, we randomize the indexes of agents, and then partition them into either the trained population $f$ or the fictitious population $\tilde{g}$ based on the vector $\bm{v}$ drawn from the distribution $\mathfrak{B} (\bm{v}_{<i}; p, N)$. The trained population employs the learning policy networks $\bm{\theta}$, whereas the fictitious population uses the distilled policy networks $\bm{\phi}$. 
% To be clear, we rephrase $f$ and $\tilde{g}$ as $f_{\bm{\theta}}$ and $\tilde{g}_{\bm{\phi}}$ respectively. 

\emph{Line 7:} We then deploy the trained and fictitious populations to play within the substrate and generate a diverse set of multi-agent trajectories, which aim to emulate those generated by interacting with the background population in the testing scenario. 

\emph{Line 8:} To optimize the learning policies $\bm{\theta}$, we instantiate our BiDist on the MAPPO algorithm\footnote{BiDist can be instantiated with any MARL algorithms by sampling policies from $\tilde{h}_i$.}~\cite{yu2021mappo}, which is an effective on-policy MARL algorithm grounded in the CTDE paradigm \cite{oliehoek2008optimal}. The clip objective can be formally defined as: 
\begin{align}
    % \label{eq:per_capita_return_test}
    &\mathcal{J}(\bm{\theta})=\frac{1}{N}\sum_{i=1}^N \mathbb{E}_{\pi_1^{\rm old}\sim \tilde{h}_1,...,\pi_N^{\rm old}\sim \tilde{h}_N} \mathbb{E}_{\{s,o_i,a_i\}\sim \mathcal{D}}\bigg[\nonumber\\
    &\qquad\quad\min\left( \frac{\pi_{\theta_i}(a_i|o_i)}{\pi_i^{\rm old}(a_i|o_i)} A_i,{\rm clip}\left( \frac{\pi_{\theta_i}(a_i|o_i)}{\pi_i^{\rm old}(a_i|o_i)} ,1\pm\epsilon\right)A_i \right) \bigg],
\end{align}
where $\tilde{h}_i=f^{v_i} \cdot \tilde{g}^{1-v_i}$ and $\{\pi_i^{\rm old}\}_{i=1}^N$ are the behavior policies that are used to collect the trajectories in Line 7.
$\hat{A}^{i}$ is a generalized advantage estimator~\cite{schulman2015gae} of the joint advantage function.

\emph{Lines 9-13:} The distilled policies' parameters $\bm{\phi}$ are alternately updated by forward distillation and reverse distillation every $k_d$ iterations, forming a mixed-play framework. During the forward phase, the fictitious population follows the historical policies' space and forms an implicit self-play. On the other hand, during the reverse phase, the fictitious population moves away from the historical policies' space and forms a non-self-play. 

%% file: body/theoretical.tex
Next, we provide a theoretical analysis and demonstrate how our BiDist method aids in generalization and gain a deeper conceptual understanding of why our model effectively enhances generalization. All the proofs pertinent to this section are included in Appendix~A. 
% Appendix~\ref{appendix:proofs}. 
We begin by introducing several key definitions.
\begin{definition}
    \emph{We define $\Phi$ as an an arbitrary MARL algorithm, $\mathbb{P}$ and $\mathbb{P}_{\mathcal{Z}^{\prime}}$ as the training and testing distributions respectively, $\Phi_\mathbb{P}$ as a MARL algorithm under training distribution $\mathbb{P}$, and the loss function $l(o_i,a_i^{\star};\Phi_\mathbb{P})$ as the 2-norm between the softmax output of policy $\pi_i(\cdot|o_i)$ and the ground-truth optimal action $a_i^{\star}$ of the zero-shot task.}
\end{definition}
The 2-norm loss $l(o_i, a_i^{\star}; \Phi_\mathbb{P})$ measures the degree to which the output of the policy $\pi_i$ aligns with the ground-truth optimal action. A smaller value of $l(o_i, a_i^{\star}; \Phi_\mathbb{P})$ indicates that $\Phi_\mathbb{P}$ produces a policy that more closely approximates the optimal action.
We then demonstrate that the policy function $\pi_i$ and the loss function $l(o_i, a_i^{\star}; \Phi_\mathbb{P})$ are Lipschitz continuous, which forms an important foundation for the final theorem. We thus present the following lemma:
\begin{lemma}
\label{lemma:Lipschitz}
    If $\pi_i$ is parameterized by a neural network with $n_c$ convolutional layers, $n_a$ attention layers and $n_f$ fully connected layers, the policy function $\pi_i$ and loss function $l(o_i,a_i^{\star};\Phi_\mathbb{P})$ are $\lambda$-Lipschitz continuous with Lipschitz constant $\lambda=\frac{\sqrt{|\mathcal{A}_i|-1}}{|\mathcal{A}_i|} \alpha^{n_c + n_f}\left(\alpha^3\beta^2+\alpha\right)^{n_a}$.
\end{lemma}
It holds for a wide range of network architectures, including convolutional, attention-based, and fully connected networks, thereby offering strong practical significance.
With Lemma~\ref{lemma:Lipschitz}, we leverage the concept of $\delta$-cover~\cite{sammut2017encyclopedia,sener2018active} and introduce a theorem that elucidates the relationship between $\delta$ and generalization error bound.
\begin{theorem}
    \label{theorem:generalization_error_bound}
    Given that the policy function $\pi_i$ and loss function $l(o_i,a_i^{\star};\Phi_\mathbb{P})$ are $\lambda$-Lipschitz continuous, loss function is bounded by $L$, $\mathbb{P}$ is $\delta$ cover of $\mathbb{P}_{\mathcal{Z}^{\prime}}$, and $l(o_i,a_i^{\star};\Phi_\mathbb{P})=0$ for $o_i\in\mathbb{P}$, with the probability at least $1-\gamma$, the generalization error bound satisfies:
    $$
    \mathbb{E}_{o_i,a_i^{\star}\sim\mathbb{P}_{\mathcal{Z}^{\prime}}}[l(o_i,a_i^{\star};\Phi_\mathbb{P})]\leq\delta\lambda(1+ L|\mathcal{A}_i|)+\sqrt{\frac{L^2\log(1/\gamma)}{2n}}.\nonumber
    $$
\end{theorem}
Theorem~\ref{theorem:generalization_error_bound} reveals that $\delta$ is positively correlated with the generalization error bound.
In this context, the statement ``$\mathbb{P}$ is a $\delta$-cover of $\mathbb{P}_{\mathcal{Z}^{\prime}}$'' means a set of balls with radius $\delta$ centered at each sample of $\mathbb{P}$ can cover the entire testing distribution $\mathbb{P}_{\mathcal{Z}^{\prime}}$. 

As shown in Figure~\ref{fig:inside-outside}, the standard MARL distribution requires a large radius $\delta$ to adequately cover both the inside-space and outside-space regions of the testing distribution. In contrast, self-play (or forward distillation) effectively expands the training distribution, enabling it to nearly cover the entire inside-space distribution with $\delta \to 0$. Since a smaller $\delta$ correlates with a reduced error bound, self-play proves highly effective in addressing the inside-space generalization problem. However, a larger radius $\delta$ is still required to cover the entire testing distribution. Furthermore, merely increasing policy interactions within the self-play distribution does little on the generalization error bound unless the covering radius is sufficiently reduced. BiDist overcomes this limitation through preference shifts introduced via reverse distillation,  which diversifies the policy beyond the confines of the self-play distribution. This process further reduces $\delta$, leading to a more robust generalization to the outside-space distribution.

%% file: body/experiment.tex
In this section, we present the empirical evaluation of BiDist across various population-population generalization tasks. Our experiments are designed to demonstrate \textbf{(1)} its generalization ability to handle unseen population across cooperation, competition, and social dilemma games; \textbf{(2)} its effectiveness in diversifying the outside-space distribution space; \textbf{(3)} the effectiveness of KL divergence maximization in reverse phase in inducing preference shift; \textbf{(4)} the ablation studies that show the contribution of each component.

\subsection{Experimental Setup}
\begin{figure}[t]
    \centering
    \includegraphics[width=1\columnwidth]{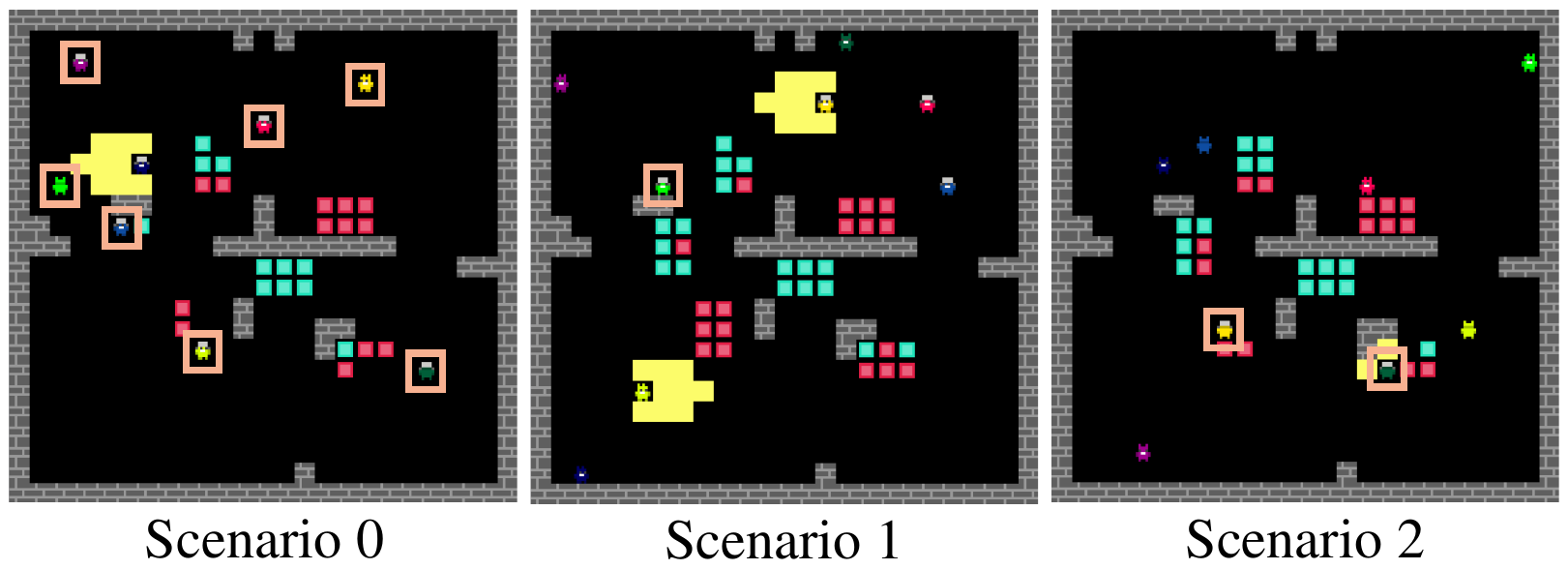}
    \caption{Illustrative instances of the testing scenarios (0-2) for the substrate \emph{Prisoners Dilemma in the matrix: Arena}. The background population is visually emphasized through bounding boxes. Pre-trained background population varies in different scenarios, including the number of agents and decision-making preferences.}
    \label{fig:demo-sc0-2}
\end{figure}

\paragraph{Generalization Task.}
In our experiment, we carry out the following 5 different generalization tasks, including cooperation, competition, and social dilemmas, on DeepMind's Melting Pot~\cite{agapiou2022melting}. 
\begin{itemize}
    \item \emph{\textbf{Pure Coordination} in the matrix: Repeated.} Players need to collaborate and coordinate to collect resources of the same color to achieve consensus and receive rewards.
    \item \emph{\textbf{Coop Mining}.} Players need to cooperate in mining gold ore, but they can also individually mine iron ore. Cooperating to mine gold ore yields higher rewards.
    \item \emph{\textbf{Chicken} in the matrix: Repeated.} Players adopt 'hawk' or 'dove' strategies while collecting resources of different colors, attempting to exploit each other, which may lead to suboptimal outcomes for both.
    \item \emph{\textbf{Coins}.} Two players collect coins of different colors, and matching colors results in negative rewards, requiring a careful balancing of collection strategies.
    \item \emph{\textbf{Prisoners Dilemma} in the matrix: Arena.} Players encounter the prisoner's dilemma while collecting resources, needing to balance the conflict between individual and group rewards.
\end{itemize}
Melting Pot is a comprehensive test suite designed to evaluate the generalization ability of MARL algorithms. It provides a rigorous and interpretable evaluation metric: during the training phase, all agents are trained together in a substrate; in the subsequent testing scenarios, some agents are replaced by pre-trained background populations that are exclusively used for testing purposes, and the remaining trained agents (focal population) are used to assess the generalization performance. 
Figure~\ref{fig:demo-sc0-2} illustrates the association of every single training substrate with multiple testing scenarios, each involving different pre-trained background populations of agents to simulate distinct social dynamics.

\paragraph{Baselines.}
Our baselines include MAPPO~\cite{yu2021mappo}, RandNet~\cite{lee2019network}, OPRE~\cite{vezhnevets2020options}, population-based self-play (PP), and RPM~\cite{qiu2022rpm}. MAPPO, an extension of the well-established PPO algorithm~\cite{schulman2017proximal}, represents the current state-of-the-art in MARL algorithms. RandNet introduces a randomized layer to improve the robustness of individual agents in previously unseen environments. OPRE introduces a hierarchical agent architecture aimed at enhancing the generalization of MARL agents. PP extends the naive self-play~\cite{silver2018general} to population-level interactions, where agents randomly sample policies from a pool of all current and historical policies. 
RPM is the state-of-the-art self-play framework designed for population-population generalization in MARL. It introduces a ranked policy pool and a hierarchical sampling mechanism to optimize generalization performance.

\paragraph{Training details.}
To ensure fairness, all algorithms utilize the same network architecture as described in~\cite{qiu2022rpm}. Parameter sharing is adopted, with all agents sharing the same actor, critic, and distilled networks. Default hyperparameters from the original papers are retained for all baselines to maintain their optimal performance. The distillation interval $k_d$ of BiDist is set to 5, ensuring a temporal lag between distilled and learning policies. The probability parameter $p$ takes 0.4 for tasks involving fewer agents and 0.2 for those with a larger agent count. The learning rate $\eta_f$ for the forward phase is $1\times10^{-3}$, while a smaller learning rate $\eta_r=1\times10^{-5}$ for the reverse phase is used to avoid the emergence of infeasible or ineffective distilled policies. More details are provided in Appendix~B. 
\begin{table*}[htbp]
\centering
\caption{The normalized focal per-capita returns of different algorithms on the testing scenarios for each substrate. SC denotes the scenario. The returns are min-max normalized within each scenario and therefore 0.0 and 1.0 represent the worst and the best respectively.}
\label{tab:main}
\begin{tabular}{@{\,}l@{\,}|@{\,}c@{\,\,}c@{\,\,}c@{\,\,}c@{\,\,}c@{\,}|@{\,}c@{\,\,}c@{\,\,}c@{\,\,}c@{\,\,}c@{\,}|@{\,}c@{\,\,}c@{\,\,}c@{\,\,}c@{\,\,}c@{\,}|@{\,}c@{\,\,}c@{\,\,}c@{\,\,}c@{\,\,}c@{\,}|@{\,}c@{\,\,}c@{\,\,}c@{\,\,}c@{\,\,}c@{\,}}
\toprule
 & \multicolumn{5}{@{\,}c@{\,}|@{\,}}{Pure Coordination} & \multicolumn{5}{@{\,\,}c@{\,}|@{\,}}{Coop Mining} & \multicolumn{5}{@{\,\,}c@{\,}|@{\,}}{Chicken} & \multicolumn{5}{@{\,\,}c@{\,}|@{\,}}{Coins} & \multicolumn{5}{@{\,\,}c@{\,}}{Prisoners Dilemma} \\
\midrule
Method & SC0 & SC1 & SC2 & SC3 & SC4 & SC0 & SC1 & SC2 & SC3 & SC4 & SC0 & SC1 & SC2 & SC3 & SC4 & SC0 & SC1 & SC2 & SC3 & SC4 & SC0 & SC1 & SC2 & SC3 & SC4 \\
\midrule
MAPPO & 0.50 & 0.61 & 0.41 & 0.67 & 0.32 & 0.40 & 0.50 & 0.00 & 0.61 & 0.79 & 0.66 & 0.60 & 0.72 & 0.71 & 0.71 & 0.97 & 0.87 & 0.87 & 0.97 & 0.73 & 0.64 & 0.60 & 0.75 & 0.88 & 0.89 \\
RanNet & 0.00 & 0.00 & 0.00 & 0.00 & 0.00 & 0.00 & 0.00 & 0.11 & 0.00 & 0.00 & 0.00 & 0.00 & 0.00 & 0.00 & 0.00 & 0.00 & 0.00 & 0.00 & 0.00 & 0.00 & 0.17 & 0.00 & 0.00 & 0.00 & 0.00 \\
OPRE & 0.40 & 0.58 & 0.26 & 0.86 & 0.03 & 0.52 & 0.74 & 0.93 & 0.78 & 0.72 & 0.54 & 0.62 & 0.61 & 0.72 & 0.72 & 0.93 & 0.92 & 0.75 & \textbf{1.00} & \textbf{1.00} & 0.00 & 0.27 & 0.14 & 0.20 & 0.29 \\
PP & 0.65 & 0.43 & 0.81 & 0.72 & 0.64 & 0.25 & 0.26 & 0.22 & 0.18 & 0.04 & 0.47 & 0.43 & 0.36 & 0.53 & 0.59 & 0.86 & 0.64 & 0.76 & 0.21 & 0.75 & 0.66 & 0.92 & 0.63 & 0.77 & 0.52 \\
RPM & 0.77 & 0.65 & 0.79 & 0.80 & 0.71 & 0.18 & 0.29 & 0.37 & 0.02 & 0.05 & 0.52 & 0.52 & 0.48 & 0.48 & 0.49 & 0.88 & 0.81 & 0.72 & 0.09 & 0.80 & 0.80 & \textbf{1.00} & 0.79 & 0.56 & 0.68 \\
\midrule
BiDist & \textbf{\textbf{1.00}} & \textbf{1.00} & \textbf{1.00} & \textbf{1.00} & \textbf{1.00} & \textbf{1.00} & \textbf{1.00} & \textbf{1.00} & \textbf{1.00} & \textbf{1.00} & \textbf{1.00} & \textbf{1.00} & \textbf{1.00} & \textbf{1.00} & \textbf{1.00} & \textbf{1.00} & \textbf{1.00} & \textbf{1.00} & 0.97 & 0.99 & \textbf{1.00} & 0.96 & \textbf{1.00} & \textbf{1.00} & \textbf{1.00} \\
\bottomrule
\end{tabular}
\end{table*}

\subsection{Generalization Performance}
In this part, we evaluate the population-population generalization ability of different algorithms across cooperation, competition, and social dilemma games. The results are presented in Table~\ref{tab:main}. We report the min-max normalized focal per-capita returns of the first five test scenarios (scenarios 0-4) for each substrate. 

We observe that the performance of RanNet, which employs a data augmentation-like technique to bolster individual agent robustness, falls short of expectations. It may not learn meaningful or effective policies for complex social tasks, therefore struggling to achieve population-population generalization. Furthermore, its inferior performance relative to MAPPO suggests that the introduction of a single layer of randomness could exacerbate the complexity of training in unseen co-player generalization tasks. In comparison, OPRE, PP, and RPM achieve competitive outcomes in certain tasks. However, their generalization ability displays discontinuity in different scenarios. Notably, both algorithms exhibit performance fluctuations across different scenarios within a single substrate. For example, RPM ranges from a peak of 1.00 in the prisoner's dilemma to a low of 0.56, whereas OPRE varies from 0.86 in pure coordination scenarios to a mere 0.03 at its weakest. These inconsistencies highlight their difficulties in effectively incorporating preference shifts to address the outside-space generalization. Consequently, while their performance may shine in certain scenarios, they falter in others where agents hold disparate preferences. 

In contrast, our BiDist consistently achieves high normalized focal per-capita returns across varying dynamics. By integrating preference shifts into the training data, BiDist more effectively addresses both inside-space and outside-space challenges.

\begin{figure}[t]
    \centering
    \includegraphics[width=1\columnwidth]{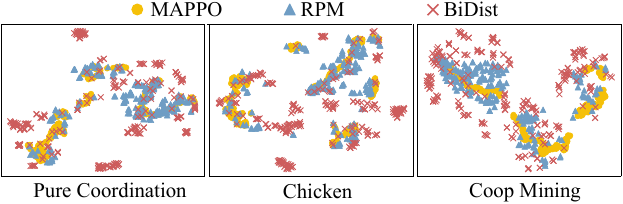}
    \caption{The t-SNE results of different algorithms on \emph{Pure Coordination}, \emph{Chicken}, and \emph{Coop Mining} tasks. Each sample represents the joint probability distribution of all agents' actions.}
    \label{fig:demo-sc0-tsne}
\end{figure}
\subsection{Diversified Distribution}
Next, we demonstrate that BiDist can generate more extensive interactions and diversify the outside-space distribution space. To achieve this, we employ the t-SNE~\cite{van2008visualizing} technique to visualize the joint probability distributions of agents' actions.

As illustrated in Figure~\ref{fig:demo-sc0-tsne}, the policy distribution generated by the MAPPO are the most compact, forming several tightly concentrated clusters. This indicates that MAPPO primarily operates within a confined distribution space, as it only facilitates coordination among agents’ \emph{current} policies. In contrast, the RPM method displays a wider spread in distribution. However, this distribution is still relatively limited to interactions within the inside-space, as RPM depends on self-play and historical policy reuse. Consequently, the blue samples are diverged only around the yellow samples. This limitation prevents self-play from adequately addressing outside-space challenges, as indicated in Figure~\ref{fig:inside-outside} and from further reducing the generalization error bound, as discussed in Theorem~\ref{theorem:generalization_error_bound}.

Our BiDist, on the other hand, exhibits a significantly broader and more diverse distribution. Notably, BiDist not only captures the distributional patterns of self-play but also extends well beyond these boundaries. For example, the red samples effectively cover the regions occupied by the self-play distribution (blue samples), while simultaneously forming distinct clusters outside these areas. This process further reduces $\delta$ in Theorem~\ref{theorem:generalization_error_bound}, enhancing the model’s ability to generalize to outside-space distributions.

\subsection{Perturbation in Reverse Phase}
\begin{figure}[t]
    \centering
    \includegraphics[width=1\columnwidth]{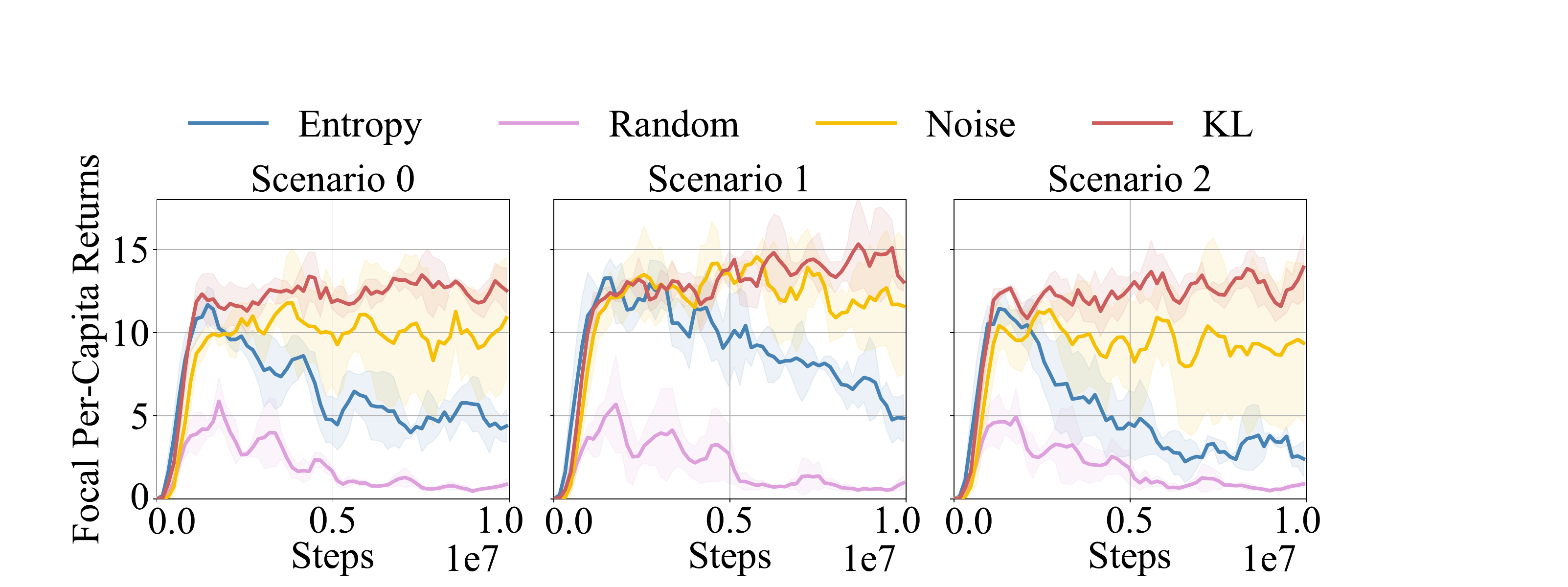}
    \caption{Performance comparison of different perturbations on \emph{Pure Coordination} task. Entropy, Random, Noise, and KL for entropy maximization, randomization, noise injection, and KL divergence maximization respectively.}
    \label{fig:reverse_type}
\end{figure}
Now, we analyze whether the KL divergence maximization in reverse distillation exhibits meaningful abstractions. To do so, we replace it with other perturbation types, including entropy maximization, randomization, and noise injection. 

\begin{figure}[t]
\centering
\includegraphics[width=1\columnwidth]{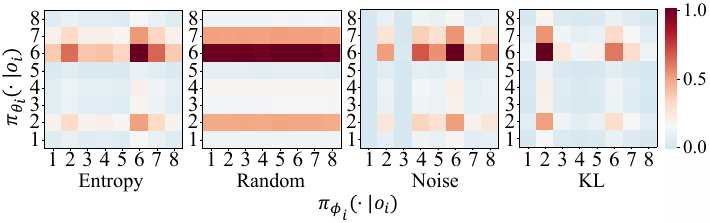} 
\caption{Normalized joint probability distribution of distilled policy and learning policy under different perturbations. The horizontal and vertical axes correspond to $\pi_{\phi_i}(\cdot|o_i)$ and $\pi_{\theta_i}(\cdot|o_i)$ respectively. $1-8$ denote the actions $1-8$. Hence, each $8\times8$ diagram illustrates the joint probabilities $\pi_{\phi_i}(\cdot|o_i)\times\pi_{\theta_i}(\cdot|o_i)$ in a min-max normalized manner.}
\label{fig:heatmap}
\end{figure}

The results presented in Figure~\ref{fig:reverse_type} demonstrate that the KL divergence maximization yields superior performance, underscoring its efficacy as the most potent perturbation type in reverse distillation. In order to validate Proposition~\ref{proposition:preference_shift} and to understand how different perturbation types affect the quality of the distilled policies, we visualize joint probability distribution (normalized) of the learning policy $\pi_{\theta_i}(\cdot|o_i)$ and the distilled policy $\pi_{\phi_i}(\cdot|o_i)$ under different perturbations, as illustrated in Figure~\ref{fig:heatmap}. Notably, we observe that the preferred action of $\pi_{\theta_i}$ is action 6. However, among the perturbation strategies considered (entropy maximization, randomization, and noise injection), none of them succeed in inducing preference shifts. Particularly, the randomization that leads to a uniform distribution is the worst case, as it degrades the quality of the distilled policies or makes them irrelevant to the task. It is worth highlighting that while noise injection does not achieve preference shifts, it enhances the likelihood of actions 4 and 5, which can be interpreted as a partial shift. This contributes to improved performances compared to entropy maximization and randomization. However, due to the inherent randomness of noise injection, the predictability of preference shifts remains highly elusive. Ultimately, the KL divergence maximization achieves the desired preference shift from action 6 of $\pi_{\theta_i}$ to action 2 of $\pi_{\phi_i}$, enabling the fictitious population to comprehensively encompass a spectrum of action preferences that may arise from outside-space distribution.

\subsection{Ablations}
    \begin{figure}[t]
        \centering
        \includegraphics[width=1\columnwidth]{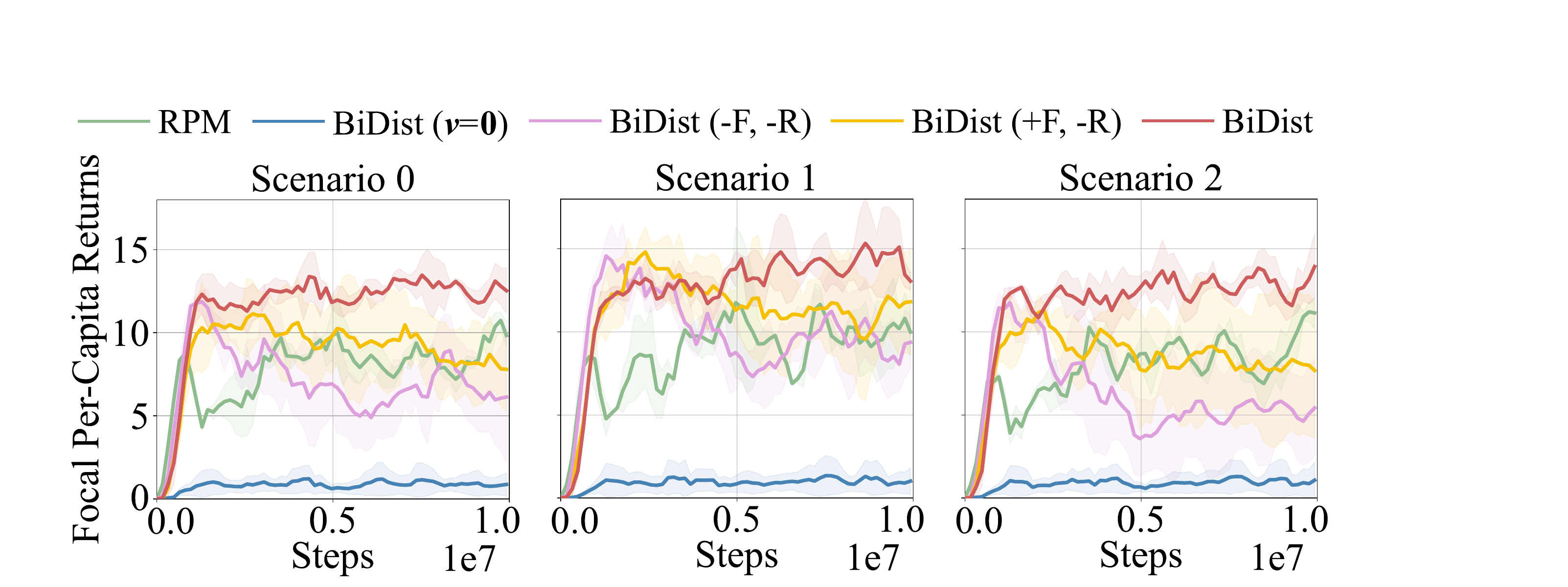} 
        \caption{Ablation studies of BiDist on \emph{Pure Coordination} task. RPM is for comparison.}
        \label{fig:ablation}
    \end{figure}
We finally evaluate the ablations of BiDist to demonstrate the effectiveness of each component. Specifically, we examine the impact of the fictitious population by setting $\bm{v}=\{0\}_{\times N}$ (\emph{BiDist ($\bm{v}=\bm{0}$)}), and then we investigate the significance of reverse distillation by excluding it from the model (\emph{BiDist (+F, -R)}). Building upon this, we take a step further by eliminating the forward distillation to assess its role in the generalizable performance (\emph{BiDist (-F, -R)}).

As indicated in Figure~\ref{fig:ablation}, a significant performance deterioration is observed in the case of BiDist ($\bm{v}=\bm{0}$). This finding verifies the meaninglessness of enabling all agents to employ distilled policies, as scenarios where all agents act as background agents are nonexistent. Moreover, we observe a substantial decline in the performance of BiDist (+F, -R) and BiDist (-F, -R), indicating the indispensability of both forward and reverse distillations. Intriguingly, forward distillation alone achieves a compatible performance (but slightly overfitting) with the self-play-based RPM. It is intuitive that distillation is unable to extract 100\% knowledge of the historical policies due to inherent deviations. However, these deviations could, in turn, enrich the decision-making patterns of distilled networks outside of the self-play, thus bolstering the performance. These insights demonstrate that the generalization of MARL agents doesn't solely depend on self-play policies but also relies on valuable multi-agent interactions beyond self-play. Additionally, forward alone (BiDist (+F, -R)) is not sufficient for addressing complex scenarios where the background population may exhibit preference shifts. In such contexts, reverse distillation proves essential for generating innovative policies for the fictitious population, enabling them to navigate diverse social situations.

%% file: body/conclusion.tex
In this paper, we addressed the challenge of population-population generalization in MARL and proposed a concise and effective approach, called BiDist. Central to our contribution is the formulation of a mixed-play framework that leverages the power of diversity in agent interactions. Our framework randomly detaches a subset of agents from the trained population and uses BiDist to generate diverse policies for them. BiDist consists of two alternating phases: forward and reverse distillations. It presents a robust and practical structure solution for both inside-space and outside-space challenges, and effectively achieves historical knowledge retention and preference shifts for the fictitious population. Theoretical analysis and empirical results both provide support for the superiority of BiDist in the robustness and capacity for effective generalization. BiDist harnesses the power of agent-agent diversity through mixed-play, providing a promising paradigm for advancing the generalization in MARL. Hence, one direction for future work is dynamically balancing exploration and exploitation in mixed-play for the specific task to enhance training efficiency. Another promising idea lies in long-term knowledge retention and adaptation, for example, through continuous learning~\cite{kirkpatrick2017overcoming}.

%% file: body/appendix.tex
\renewcommand{\theequation}{\thesection.\arabic{equation}}
\renewcommand{\thefigure}{\thesection.\arabic{figure}}
\renewcommand{\thetable}{\thesection.\arabic{table}}
\makeatletter
\@addtoreset{equation}{section}
\@addtoreset{figure}{section}
\@addtoreset{table}{section}
\makeatother
\onecolumn
\section{Proofs}
\label{appendix:proofs}
\subsection{Proof of Proposition~\ref{proposition:preference_shift}}
\begin{proof}
    Let $N_a$ denote the action space size. We start by presenting the KL divergence between $\pi_{\theta_i}$ and $\pi_{\phi_i}$:
    \begin{equation}
        D_{\rm KL}\left(\pi_{\theta_i}(\cdot|o_i) || \pi_{\phi_i}(\cdot|o_i)\right)=\sum_{j=1}^{N_a}{\pi_{\theta_i}(a_i^j|o_i)}\cdot\log\frac{\pi_{\theta_i}(a_i^j|o_i)}{\pi_{\phi_i}(a_i^j|o_i)}.
    \end{equation}
    Let $\pi_{\tilde{\phi}_i}(\cdot|o_i)$ denote the updated distilled policy of the reverse distillation. Incorporating update bias, we represent $D_{\rm KL}$ after the update as $\sum_{j}\pi_{\theta_i}(a_i^j|o_i)\cdot\log\frac{\pi_{\theta_i}(a_i^j|o_i)}{\pi_{\phi_i}(a_i^j|o_i)+\Delta_j}$, where $\Delta_j$ is the update bias for each action $j$, with $\sum_j^{N_a}\Delta_j=0$. Thus, the change of $D_{\rm KL}$ before and after update can be given by
    \begin{align}
        &D_{\rm KL}\big(\pi_{\theta_i}(\cdot|o_i) || \pi_{\tilde{\phi}_i}(\cdot|o_i)\big)-D_{\rm KL}\big(\pi_{\theta_i}(\cdot|o_i) || \pi_{\phi_i}(\cdot|o_i)\big)\nonumber\\
        =&\sum_{j}\pi_{\theta_i}(a_i^j|o_i)\cdot\log\frac{\pi_{\theta_i}(a_i^j|o_i)}{\pi_{\phi_i}(a_i^j|o_i)+\Delta_j}-\sum_{j}{\pi_{\theta_i}(a_i^j|o_i)}\cdot\log\frac{\pi_{\theta_i}(a_i^j|o_i)}{\pi_{\phi_i}(a_i^j|o_i)}\nonumber\\
        =&\sum_{j}\pi_{\theta_i}(a_i^j|o_i)\cdot\log\frac{\pi_{\phi_i}(a_i^j|o_i)}{\pi_{\phi_i}(a_i^j|o_i)+\Delta_j}\nonumber\\
        =&\pi_{\theta_i}(a^{\text{prefer}}_{\theta_i}|o_i)\cdot\log\frac{\pi_{\phi_i}(a^{\text{prefer}}_{\theta_i} |o_i)}{\pi_{\phi_i}(a^{\text{prefer}}_{\theta_i}|o_i)+\Delta_p} + \pi_{\theta_i}(a_i^{k}|o_i)\cdot\log\frac{\pi_{\phi_i}(a_i^{k} |o_i)}{\pi_{\phi_i}(a_i^{k}|o_i)+\Delta_k}+ \sum_{j\neq p,k}\pi_{\theta_i}(a_i^j|o_i)\cdot\log\frac{\pi_{\phi_i}(a_i^j|o_i)}{\pi_{\phi_i}(a_i^j|o_i)+\Delta_j},
    \end{align}
    where specific indices $p$ (preferred action) and $k$ (any other non-preferred action) are considered.

    To prove the existence of update margins $\{\Delta_j\}_{j\sim\{1,\dots,N_a\}}$ that drive preference shifts, we can turn to prove the existence of update margins $\Delta_p=-\Delta<0$ and $\Delta_k=\Delta>0$ while considering $\Delta_{j\neq p,k}=0$. Thus, the above equation is
    \begin{align}
        =&\pi_{\theta_i}(a^{\text{prefer}}_{\theta_i}|o_i)\cdot\log\frac{\pi_{\phi_i}(a^{\text{prefer}}_{\theta_i} |o_i)}{\pi_{\phi_i}(a^{\text{prefer}}_{\theta_i}|o_i)-\Delta} + \pi_{\theta_i}(a_i^{k}|o_i)\cdot\log\frac{\pi_{\phi_i}(a_i^{k} |o_i)}{\pi_{\phi_i}(a_i^{k}|o_i)+\Delta}+ \sum_{j\neq p,k}\pi_{\theta_i}(a_i^j|o_i)\cdot\log\frac{\pi_{\phi_i}(a_i^j|o_i)}{\pi_{\phi_i}(a_i^j|o_i)}\nonumber\\
        \geq &\pi_{\theta_i}(a^{\text{prefer}}_{\theta_i}|o_i)\cdot\log\frac{\pi_{\phi_i}(a^{\text{prefer}}_{\theta_i} |o_i)}{\pi_{\phi_i}(a^{\text{prefer}}_{\theta_i}|o_i)-\Delta} + \pi_{\theta_i}(a^{\text{prefer}}_{\theta_i}|o_i)\cdot\log\frac{\pi_{\phi_i}(a_i^{k} |o_i)}{\pi_{\phi_i}(a_i^{k}|o_i)+\Delta}\nonumber\\
        =&\pi_{\theta_i}(a^{\text{prefer}}_{\theta_i}|o_i)\cdot\log\frac{\pi_{\phi_i}(a^{\text{prefer}}_{\theta_i} |o_i)\pi_{\phi_i}(a_i^{k} |o_i)}{\pi_{\phi_i}(a^{\text{prefer}}_{\theta_i} |o_i)\pi_{\phi_i}(a_i^{k} |o_i)+\Delta\cdot \left(\pi_{\phi_i}(a^{\text{prefer}}_{\theta_i} |o_i)-\pi_{\phi_i}(a_i^{k} |o_i)-\Delta\right)}. \label{eq:change_of_KL}
    \end{align}
    Generally, on the premise of increasing KL divergence (Equation~\eqref{eq:change_of_KL} $>0$), there exists a $\Delta>\pi_{\phi_i}(a^{\text{prefer}}_{\theta_i} |o_i)-\pi_{\phi_i}(a_i^{k} |o_i)$ that enables $\pi_{\tilde{\phi}_i}(a^{\text{prefer}}_{\theta_i} |o_i)=\pi_{\phi_i}(a^{\text{prefer}}_{\theta_i} |o_i)-\Delta<\pi_{{\phi}_i}(a_i^{k} |o_i)+\Delta=\pi_{\tilde{\phi}_i}(a_i^{k} |o_i)$. This implies that $a^{\text{prefer}}_{\theta_i}$ is no longer the preferred action of distilled policy $\pi_{\tilde{\phi}_i}$, resulting in a preference shift. 

    To relate this to entropy maximization, we start from the upper bound of entropy maximization. Thus, given the distilled policy $\pi_{\phi_i}$ that reaches the upper bound $\mathcal{H}^{\max}(\pi_{\phi_i}(\cdot|o_i))$ of entropy, we have
    \begin{align}
        &\pi_{\theta_i}(a^{\text{prefer}}_{\theta_i}|o_i)\cdot\log\frac{\pi_{\phi_i}(a^{\text{prefer}}_{\theta_i} |o_i)\pi_{\phi_i}(a_i^{k} |o_i)}{\pi_{\phi_i}(a^{\text{prefer}}_{\theta_i} |o_i)\pi_{\phi_i}(a_i^{k} |o_i)+\Delta\cdot \left(\pi_{\phi_i}(a^{\text{prefer}}_{\theta_i} |o_i)-\pi_{\phi_i}(a_i^{k} |o_i)-\Delta\right)}\nonumber\\
        =&\pi_{\theta_i}(a^{\text{prefer}}_{\theta_i}|o_i)\cdot\log\frac{1}{1-(n\Delta)^2}.\label{eq:entropy_change_of_KL}
    \end{align}
    Clearly, any $\Delta>0$ results in $\pi_{\tilde{\phi}_i}(a^{\text{prefer}}_{\theta_i} |o_i)$ continuing decreasing while following the increase of KL divergence (Equation~\eqref{eq:entropy_change_of_KL} $>0$). In this case, $\pi_{\tilde{\phi}_i}(a^{\text{prefer}}_{\theta_i} |o_i)=\frac{1}{n}-\Delta$ is less than $\frac{1}{n}$, thus leading to the preference shift $a^{\text{prefer}}_{\theta_i}\neq a^{\text{prefer}}_{\phi_i}$.
\end{proof}

\subsection{Proof of Lemma~\ref{lemma:Lipschitz}}
\begin{proof}
We will first show that the softmax function defined over $\mathcal{A}_i$ action space is $\frac{\sqrt{|\mathcal{A}_i|-1}}{|\mathcal{A}_i|}$-Lipschitz continuous.

Let \(f(x)\) be the softmax function:
\begin{align}
f(x)_k = \frac{\exp(x_k)}{\sum_{j=1}^{|\mathcal{A}_i|} \exp(x_j)}, \quad k = 1, 2, \dots, |\mathcal{A}_i|.
\end{align}
The Jacobian matrix of the softmax function is given by:
\begin{align}
J =
\begin{bmatrix}
f_1(1 - f_1) & -f_1 f_2 & \cdots & -f_1 f_{|\mathcal{A}_i|} \\
-f_2 f_1 & f_2(1 - f_2) & \cdots & -f_2 f_{|\mathcal{A}_i|} \\
\vdots & \vdots & \ddots & \vdots \\
-f_{|\mathcal{A}_i|} f_1 & -f_{|\mathcal{A}_i|} f_2 & \cdots & f_{|\mathcal{A}_i|}(1 - f_{|\mathcal{A}_i|})
\end{bmatrix}.
\end{align}

The Frobenius norm of the Jacobian matrix is:
\begin{align}
\|J\|_F = \sqrt{\sum_{k=1}^{|\mathcal{A}_i|} \sum_{j=1, j \neq k}^{|\mathcal{A}_i|} f_k^2 f_j^2 + \sum_{k=1}^{|\mathcal{A}_i|} f_k^2 (1 - f_k)^2}.
\end{align}
It can be shown that when \(f_k = \frac{1}{|\mathcal{A}_i|}\) for all \(i\), the Frobenius norm reaches its maximum value:
\begin{align}
\|J\|_F^{*} = \frac{\sqrt{|\mathcal{A}_i| - 1}}{|\mathcal{A}_i|}.
\end{align}
Thus, the softmax function is \(\frac{\sqrt{|\mathcal{A}_i| - 1}}{|\mathcal{A}_i|}\)-Lipschitz continuous.

Next, we prove that the linear transformations in convolutional, fully connected, and attention layers are Lipschitz continuous. Consider the representation of the input at layer $d$:
\begin{equation}
    y_j^d = \sum_k w_{k,j}^d x^{d-1}_k
\end{equation}
Assuming the bounded weights $\sum_k |w_{k,j}^d| \leq \alpha$ and bounded input norm $\|x^d\|_2\leq \beta$, we can show that for any linear transformations in convolutional, fully connected or attention layer, the following holds:
\begin{equation}
\label{eq:linear_transformation_lipschitz}
\|y^d - \tilde{y}^d\|_2 \leq \alpha \|x^{d-1} - \tilde{x}^{d-1}\|_2.
\end{equation}
For the convolutional layer and the fully connected layer, there are a ReLU and a max-pooling followed by the linear transformation. Max-pooling can be rewritten as a linear transformation with one weight being 1 and others being 0. Combining max-pooling and ReLU which satisfies $|\max(0,a)-\max(0,b)| \leq |a-b|$, we get the output $x^d$ of convolutional, or fully connected layer:
\begin{equation}
\|x^d - \tilde{x}^d\|_2 \leq \|y^d - \tilde{y}^d\|_2 \leq \alpha \|x^{d-1} - \tilde{x}^{d-1}\|_2.
\end{equation}
We then consider the attention layer which is defined by $\text{softmax}\left(\frac{Q K^\top}{\sqrt{d_k}}\right) V$ with $Q = W_Q x$, $K = W_K x$, $V = W_V x$. Since we have proved that the part of a linear transformation is $\alpha$-Lipschitz continuous in Equation~\eqref{eq:linear_transformation_lipschitz}, we have $||Q^d-\tilde{Q}^d||_2\leq \alpha ||x^{d-1} - \tilde{x}^{d-1}||_2$, $||K^d-\tilde{K}^d||_2\leq \alpha ||x^{d-1} - \tilde{x}^{d-1}||_2$, and $\|V^d-\tilde{V}^d\|_2\leq \alpha ||x^{d-1} - \tilde{x}^{d-1}||_2$. Moreover, we can get $\|Q\|_2\leq\|W_Q\|_2\|x\|_2\leq\alpha\beta$, $\|K\|_2\leq\alpha\beta$. We let $S=\left(\frac{Q K^\top}{\sqrt{d_k}}\right)$ and get
\begin{align}
    S-\tilde{S}=\frac{Q K^\top - \tilde{Q}{\tilde{K}}^\top}{\sqrt{d_k}}=\frac{1}{\sqrt{d_k}} \left( (Q - \tilde{Q}) K^\top + \tilde{Q}(K - \tilde{K})^\top \right).
\end{align}
Thus, the following holds:
\begin{align}
    \|S-\tilde{S}\|_2=&\left\|\frac{1}{\sqrt{d_k}} \left( (Q - \tilde{Q}) K^\top + \tilde{Q}(K - \tilde{K})^\top \right)\right\|_2 \nonumber\\
    \leq&\frac{1}{\sqrt{d_k}} \left( \|(Q - \tilde{Q})\|_2\| K^\top \|_2 + \| \tilde{Q}\|_2\|(K - \tilde{K})^\top \|_2 \right)\nonumber\\
    =&\frac{1}{\sqrt{d_k}} \left( \|(Q - \tilde{Q})\|_2\| K \|_2 + \| \tilde{Q}\|_2\|(K - \tilde{K}) \|_2 \right) \nonumber\\
    \leq&\frac{1}{\sqrt{d_k}} \left( \alpha\|(x - \tilde{x})\|_2\cdot\alpha\beta + \alpha\beta\cdot\alpha\|(x - \tilde{x}) \|_2 \right)\nonumber \\
    =&\frac{1}{\sqrt{d_k}} \left( \|(Q - \tilde{Q})\|_2\| K \|_2 + \| \tilde{Q}\|_2\|(K - \tilde{K}) \|_2 \right) \nonumber\\
    \leq&\frac{1}{\sqrt{d_k}} \left(2\alpha^2\beta\|(x - \tilde{x})\|_2\right).
\end{align}
As we have proved that softmax function defined over $\mathcal{A}_i$ action space is \(\frac{\sqrt{|\mathcal{A}_i|-1}}{|\mathcal{A}_i|}\)-Lipschitz continuous. Thus, we have
\begin{align}
    \|\text{softmax}(S)-\text{softmax}(\tilde{S})\|_2\leq\frac{\sqrt{d_k - 1}}{d_k}\|S-\tilde{S}\|_2 \leq2\alpha^2\beta\sqrt{\frac{d_k - 1}{{d_k}^3}} \|(x - \tilde{x})\|_2.
\end{align}
Next, we multiply the softmax result by $V$
\begin{align}
    \|\text{softmax}(S)V-\text{softmax}(\tilde{S})\tilde{V}\|_2=&\left\|\left(\text{softmax}(S)-\text{softmax}(\tilde{S})\right)V+\text{softmax}(\tilde{S})\left(V-\tilde{V}\right)\right\|_2 \nonumber\\ 
    \leq&\left\|\left(\text{softmax}(S)-\text{softmax}(\tilde{S})\right)\right\|_2\left\|V\right\|_2+\left\|\text{softmax}(\tilde{S})\right\|_2\left\|\left(V-\tilde{V}\right)\right\|_2\nonumber\\
    \leq&2\alpha^2\beta\sqrt{\frac{d_k - 1}{{d_k}^3}} \|(x - \tilde{x})\|_2\cdot\alpha\beta+\alpha\|x-\tilde{x}\|_2\nonumber\\
    \leq&\left(2\alpha^3\beta^2\sqrt{\frac{d_k - 1}{{d_k}^3}}+\alpha\right)\|x-\tilde{x}\|_2, \nonumber\\
    &\text{which, to be more concise, satisfies} \nonumber\\
    \leq&\left(2\alpha^3\beta^2\frac{1}{2}+\alpha\right)\|x-\tilde{x}\|_2\nonumber\\
    =&\left(\alpha^3\beta^2+\alpha\right)\|x-\tilde{x}\|_2,
\end{align}
where $\|\text{softmax}(\tilde{S})\|_2\leq1$ is due to the fact that elements in $\text{softmax}(\tilde{S})$ are all between 0 and 1.

Therefore, we finally get the output of the attention layer satisfies
\begin{equation}
    \|x^d - \tilde{x}^d\|_2 = \|\text{Attention}(x^{d-1})-\text{Attention}(\tilde{x}^{d-1})\|_2\leq\left(\alpha^3\beta^2+\alpha\right)\|x^{d-1}-\tilde{x}^{d-1}\|_2.
\end{equation}
By combining these results, we can conclude that for a policy $\pi_i$ consisting of $n_c$ convolutional layers, $n_a$ attention layers, and $n_f$ fully connected layers, the following holds:
\begin{equation}
\| \pi_i(\cdot|o_i) - \pi_i(\cdot|\tilde{o}_i)\|_2 \leq \frac{\sqrt{|\mathcal{A}_i|-1}}{|\mathcal{A}_i|} \alpha^{n_c + n_f}\left(\alpha^3\beta^2+\alpha\right)^{n_a}\|o_i - \tilde{o_i}\|_2.
\end{equation}
Using the reverse triangle inequality, we obtain:
\begin{align}
    |l(o_i, a_i^\star; \Phi_\mathbb{P}) - l(\tilde{o_i}, a_i^\star; \Phi_\mathbb{P})| 
    &= \left\| \pi_i(\cdot|o_i) - a_i^\star \right\|_2 - \left\| \pi_i(\cdot|\tilde{o}_i) - a_i^\star \right\|_2 \nonumber\\
    &\leq \left\| \pi_i(\cdot|o_i) - \pi_i(\cdot|\tilde{o}_i) \right\|_2 \nonumber\\
    &\leq \frac{\sqrt{|\mathcal{A}_i|-1}}{|\mathcal{A}_i|} \alpha^{n_c + n_f}\left(\alpha^3\beta^2+\alpha\right)^{n_a}\|o_i - \tilde{o_i}\|_2.
\end{align}
Therefore, the loss function $l(o_i, a_i^\star; \Phi_\mathbb{P})$ is $\frac{\sqrt{|\mathcal{A}_i|-1}}{|\mathcal{A}_i|} \alpha^{n_c + n_f}\left(\alpha^3\beta^2+\alpha\right)^{n_a}$-Lipschitz continuous.
\end{proof}

\subsection{Proof of Theorem~\ref{theorem:generalization_error_bound}}
\begin{proof}
First, the following function holds:
\begin{align}
\pi_i(\cdot|o_i) = \pi_i(\cdot|o_i) + \pi_i(\cdot|o_j) - \pi_i(\cdot|o_j)
\leq \pi_i(\cdot|o_j) + |\pi_i(\cdot|o_i) - \pi_i(\cdot|o_j)|.
\end{align} 
Let $o_i\sim\mathbb{P}_{\mathcal{Z}^{\prime}}$ and $o_j\sim\mathbb{P}$. Then we can get
\begin{align}
\mathbb{E}_{a_i\sim\pi_i(\cdot|o_i)}\left[l(o_i,a_i;\Phi_\mathbb{P})\right]
    &= \sum_{ a_i \in \mathcal{A}_i} l(o_i, a_i; \Phi_P) \cdot \pi_i(a_i|o_i) \nonumber\\
    &\leq \sum_{a_i \in \mathcal{A}_i} l(o_i, a_i; \Phi_P) \cdot \Big(\pi_i(a_i|o_j) + |\pi_i(a_i|o_i) - \pi_i(a_i|o_j)|\Big)\nonumber\\
    &= \sum_{a_i \in \mathcal{A}_i} l(o_i, a_i; \Phi_P) \cdot \pi_i(a_i|o_j) + \sum_{a_i \in \mathcal{A}_i} l(o_i, a_i; \Phi_P) \cdot |\pi_i(a_i|o_i) - \pi_i(a_i|o_j)|.
\end{align}
\emph{Bounding the first term}. Given $l(o_j,a_i;\Phi_\mathbb{P})=0$ for $o_j \in \mathbb{P}$ and $l(o_i,a_i;\Phi_\mathbb{P})$ is $\lambda$-Lipschitz continuous, we have:
\begin{align}
    \sum_{a_i \in \mathcal{A}_i} l(o_i, a_i; \Phi_P) \cdot \pi_i(a_i|o_i) 
    &= \sum_{a_i \in \mathcal{A}_i} (l(o_i, a_i; \Phi_P) - l(o_j, a_i; \Phi_P)) \cdot \pi_i(a_i|o_i)\nonumber\\
    &\leq \sum_{a_i \in \mathcal{A}_i} |l(o_i, a_i; \Phi_P) - l(o_j, a_i; \Phi_P)| \cdot \pi_i(a_i|o_i)\nonumber\\
    &\leq \sum_{a_i \in \mathcal{A}_i} \lambda \cdot |o_i - o_j| \cdot \pi_i(a_i|o_i)\nonumber\\
    &\leq \lambda \delta.
\end{align}
\emph{Bounding the second term}. Given the loss function is bounded by $L$ and the policy function $\pi_i$ is $\lambda$-Lipschitz continuous, we get:
\begin{align}
    \sum_{a_i \in \mathcal{A}_i} l(o_i, a_i; \Phi_P) \cdot |\pi_i(a_i|o_i) - \pi_i(a_i|o_j)| 
    &\leq \sum_{a_i \in \mathcal{A}_i} L \cdot |\pi_i(a_i|o_i) - \pi_i(a_i|o_j)| \nonumber\\
    &\leq \sum_{a_i \in \mathcal{A}_i} L \cdot \lambda |o_i - o_j| \nonumber\\
    &\leq L \lambda \delta |\mathcal{A}_i|.
\end{align}
Combining the above results, we obtain:
\begin{align}
    \mathbb{E}_{a_i\sim\pi_i(\cdot|o_i)}[l(o_i,a_i;\Phi_\mathbb{P})] \leq \delta\lambda(1 + L|\mathcal{A}_i|).
\end{align}
For $n$ independent samples drawn from $\mathbb{P}_{\mathcal{Z}^{\prime}}$, Hoeffding's inequality gives:
\begin{align}
P \left( \mathbb{E}_{o_i,a_i^{\star} \sim \mathbb{P}_{\mathcal{Z}^{\prime}}}[l(o_i,a_i^{\star};\Phi_\mathbb{P})] - \mathbb{E}_{o_i\sim\mathbb{P}_{\mathcal{Z}^{\prime}},a_i\sim \pi_i(\cdot|o_i)}[l(o_i,a_i;\Phi_\mathbb{P})] \leq \epsilon \right) \geq 1- \exp\left( - \frac{2n^2\epsilon^2}{nL^2} \right).
\end{align}
Set $\epsilon=\sqrt{\frac{L^2\log(1/\gamma)}{2n}}$, we have, with probability at least $1 - \gamma$, the following inequality holds:
\begin{align}
\mathbb{E}_{o_i,a_i^{\star} \sim \mathbb{P}_{\mathcal{Z}^{\prime}}}[l(o_i,a_i^{\star};\Phi_\mathbb{P})] &\leq \mathbb{E}_{o_i\sim\mathbb{P}_{\mathcal{Z}^{\prime}},a_i \sim \pi_i(\cdot|o_i)}[l(o_i,a_i;\Phi_\mathbb{P})]+ \sqrt{\frac{L^2 \log(1/\gamma)}{2n}}.\nonumber\\
&\leq \delta\lambda(1 + L|\mathcal{A}_i|) + \sqrt{\frac{L^2 \log(1/\gamma)}{2n}}.
\end{align}
\end{proof}

\newpage
\section{Experimental Settings}\label{appendix:experimental_detail}

\subsection{Baselines}
\paragraph{MAPPO~\cite{yu2021mappo}.}
MAPPO is a specialized variant of PPO designed for multi-agent settings. It employs a shared policy and a centralized value function for each agent, following the CTDE paradigm. The actor network takes the local observation $o_i$ as input, while the centralized critic network takes the global state $s$. MAPPO also incorporates common practices in PPO implementations, such as advantage normalization, observation normalization, gradient clipping, and value clipping.

\paragraph{RanNet~\cite{lee2019network}.}
RandNet is proposed for enhancing the generalization of the RL agent when confronted with novel environments, particularly those featuring unfamiliar textures and layouts. The main technical implementation of RanNet is to use a random convolutional neural layer to perturb the input observations of deep reinforcement learning agents. The random layer is re-initialized at every iteration with a mixture of distributions: $P(\psi)=\alpha \mathbb{I}(\psi=\textbf{I})+(1-\alpha)\mathcal{N}(\textbf{0};\sqrt{\frac{2}{n_{\text{in}}+n_{\text{out}}}})$, where $\psi$ is the random convolutional layer, $\textbf{I}$ is an identity kernel, $\alpha\in [0,1]$ is a constant, $\mathcal{N}$ is the normal distribution, and $n_{\text{in}}$ and $n_{\text{in}}$ are the number of input and output channels. The agents are trained to learn invariant and robust features across varied and randomized environments. RanNet also proposes a feature matching (FM) loss to align the hidden features from clean and randomized inputs. Thus, the total loss is: $\mathcal{L}^{\text{random}}=\mathcal{L}^{\text{random}}_{\text{policy}}+\beta \mathcal{L}^{\text{random}}_{\text{FM}}$, where $\beta>0$ is a hyperparameter.

\paragraph{OPRE~\cite{vezhnevets2020options}.}
OPtions as REsponses (OPRE) is a hierarchical reinforcement learning approach, proposed to enhance generalization and credit assignment in multi-agent games. The aim is to ground high-level decisions in the game-theoretic structure and separate strategic and tactical knowledge into different hierarchical levels.

The key implementation of OPRE is the factorization of the policy $\pi_i$ into a shared latent variable $z$ and a mixture of components $\eta$ that represent an option:
$\pi_i(a_i | o^{\leq t}_i, s_t) = \sum_z q(z|s_t) \eta (a_i | o^{\leq t}_i, z)$, where $s_t$ is the global information. The history $o^{\leq t}_i$ is encoded via recurrent neural networks. Importantly, these ``options'' differ from options in hierarchical RL~\cite{bacon2017option}, as they lack explicit probability distributions for option entry and exit. The behavior policy is:
$\mu_i(a_i|o^{\leq t}_i) = \sum_z p(z|o_i^{\leq t}) \eta(a_i | o^{\leq t}_i, z).$ The encoder distribution $p(z|o_i^{\leq t})$ is trained by minimizing the KL divergence $D_{\rm KL}(q||p)$. In our experiments, we utilize the default 16 options for OPRE.

\paragraph{PP} Population-based self-play (PP) adopts a straightforward self-play strategy~\cite{silver2018general,baker2019emergent} by diversifying agent interactions through policy sampling from both current and historical policies. Implemented within the MAPPO framework, PP maintains a dynamic memory buffer to store policies generated at each training step.

The memory buffer contains all policies, which are classified into two categories: the current policies and a pool of historical policies. During training, agents sample policies with a 70\% probability\footnote{The probability is adopted from~\cite{qiu2022rpm}.} from the current batch and a 30\% probability from the historical pool. This mechanism ensures exposure to both current and past policies, fostering diversity in agent interactions.

\paragraph{RPM~\cite{qiu2022rpm}.}
Ranked Policy Memory (RPM) introduces a novel approach to training policies by utilizing a lookup memory called $\Psi$. The purpose of $\Psi$ is to store policies with different performance levels throughout the training process. $\Psi$ consists of multiple entries, where each entry represents a specific performance level of policies. Conceptually, $\Psi$ can be thought of as a dictionary comprising of a key $\kappa$, which determines the performance level, and a corresponding value (a list) that contains the policies at that performance level. The value of $\kappa$ is computed leveraging the training episode return $R$ of agent policies ${\{\pi_{\theta_i}\}}_{i=1}^N$, which reflects the quality of the policies. By organizing policies in $\Psi$ based on their ranks, there is a structured representation of policy performance levels.

At the beginning of each episode, agents have a probability $p$ of replacing their current policies with policies randomly sampled from $\Psi$. This mechanism allows the agents to explore and interact with diverse policies across different performance levels, leading to the generation of rich multi-agent trajectory data for training purposes. At the end of each episode, the policies are evaluated using the training episode return $R$ and subsequently appended to the corresponding entry in $\Psi$ based on their performance level.

\subsection{Networks} 
The network architectures used in our experiments were adopted from~\cite{leibo2021scalable,qiu2022rpm}. We adopt parameter sharing for all agents, meaning that all agents share the same actor network, critic network, and distilled network. We use the same network architectures for all the baseline methods and our method. Here's how these architectures were structured for the actor, critic, and distilled networks.

\paragraph{Actor network.} The actor network employs a CNN with two layers. The first layer has 16 output channels with a kernel size of 8 and a stride of 8. The second layer consists of 32 output channels with a kernel size of 4 and a stride of 1. The input to this CNN is the local observation $o_i$. Both CNN layers are activated using the ReLU activation function. Subsequent to the CNN layers, there is a two-layer MLP with 64 units each, also with ReLU activation functions. Following this, a GRU layer with 128 units is applied. The input to this GRU layer is obtained by concatenating the output of the MLP and the feature vector of each agent (e.g., position, orientation, inventory, one-hot encoding of agent ID). The output of the GRU is then fed into another MLP that generates the policy for each agent.

\paragraph{Critic network.} The critic network uses the same basic structure (2-layer CNNs, 2-layer MLP, 1-layer GRU) as the actor network with a few variations. The CNN takes the global state $s$ as input. The GRU layer takes as input the concatenation of the output from the MLP and the feature vectors of all agents (e.g., position, orientation, inventory, one-hot encoding of agent ID, agent action) as input. Subsequent to this, an additional MLP layer is employed to produce the value function for each agent.

\paragraph{Distilled network.} The architecture of the distilled network is set the same as the actor network.

\subsection{Other Details}
\paragraph{Evaluation metric.} Our evaluation metric, adapted from~\cite{wang2020rode}, involves averaging the focal per-capita return over 10 test games after each training iteration, and the final return is further averaged over 3 different training seeds.
% Our code is anonymously available at this link: \url{https://anonymous.4open.science/r/BiDist}.

\paragraph{Computing infrastructure.} We implement our BiDist with Python 3.9 and PyTorch 1.13 using GPU acceleration. All of the experiments are trained on TITAN X GPU, running 64-bit Linux 4.4.0. 

\paragraph{Hyperparameters.} 
All baselines adopt their original hyperparameters from their original papers to maintain their best performances. As for our BiDist, the hyperparameters are set the same as MAPPO paper (see Table~\ref{table:common_hyperparameter}), but with some extra hyperparameters (see Table~\ref{table:extra_hyperparameter}). The distillation interval $k_d$ is set to 5, meaning that the forward distillation and reverse distillation are alternately executed every 5 training iterations. This controlled lag ensures a deliberate temporal difference between distilled and learning policies. The probability parameter $p$ takes values of 0.2 or 0.4, contingent on the specific tasks. For tasks involving fewer agents, 0.4 is chosen, given the higher proportion of the background population. For tasks with a larger agent count, 0.2 is favored to maintain a stable training process with a controlled fictitious population size. The learning rate $\eta_f$ for the forward phase is set to $1\times10^{-3}$, while a smaller learning rate $\eta_r=1\times10^{-5}$ for the reverse phase is used to avoid the emergence of impractical or ineffective distilled policies.

\begin{table}[h]
    \centering
    \caption{Common hyperparameters between MAPPO and BiDist.}
    \label{table:common_hyperparameter}
    \begin{tabular}{c c || c c}
       \toprule
       Hyperparameters        & Value      & Hyperparameters    & Value      \\ \midrule
       Actor learning rate    & 5e-4       & Gain               & 0.01       \\
       Critic learning rate   & 5e-4       & Episode length     & 200        \\
       Gamma                  & 0.99       & Rollout threads    & 16          \\
       Gae lamda              & 0.95       & Training threads   & 32          \\
       Optimizer              & Adam       & Number of mini-batch & 5          \\
       Optimizer epsilon      & 1e-5       & Gradient clip norm & 10         \\
       Optimizer weight decay & 0          & PPO epochs         & 5       \\
       Stacked frames         & 1          & Value              & Huber loss      \\
       Entropy coef           & 0.01       & Huber delta        & 10.0       \\
       \bottomrule
    \end{tabular}
\end{table}

\begin{table}[h]
    \centering
    \caption{Extra hyperparameters of BiDist.}
    \label{table:extra_hyperparameter}
    \begin{tabular}{c c }
       \toprule
       Hyperparameters                       & Value           \\ \midrule
       Forward distillation learning rate     & 1e-3         \\
       Reverse distillation learning rate      & 1e-5 \\
       Probability $p$                        & (0.2, 0.4) \\
       Distillation internal $k_d$            & 5      \\

       \bottomrule
    \end{tabular}
\end{table}

\newpage
\section{Additional Results}
\subsection{Resource Friendly}
In this part, we explore the resource efficiency of BiDist, highlighting its advantages in this regard. One key feature of BiDist is its ability to achieve efficient resource utilization. Unlike the self-play approach that necessitates maintaining multiple rollout policy networks, BiDist only requires the upkeep of a single distilled network. 

\begin{figure}[h]
    \centering
    \includegraphics[width=0.6\textwidth]{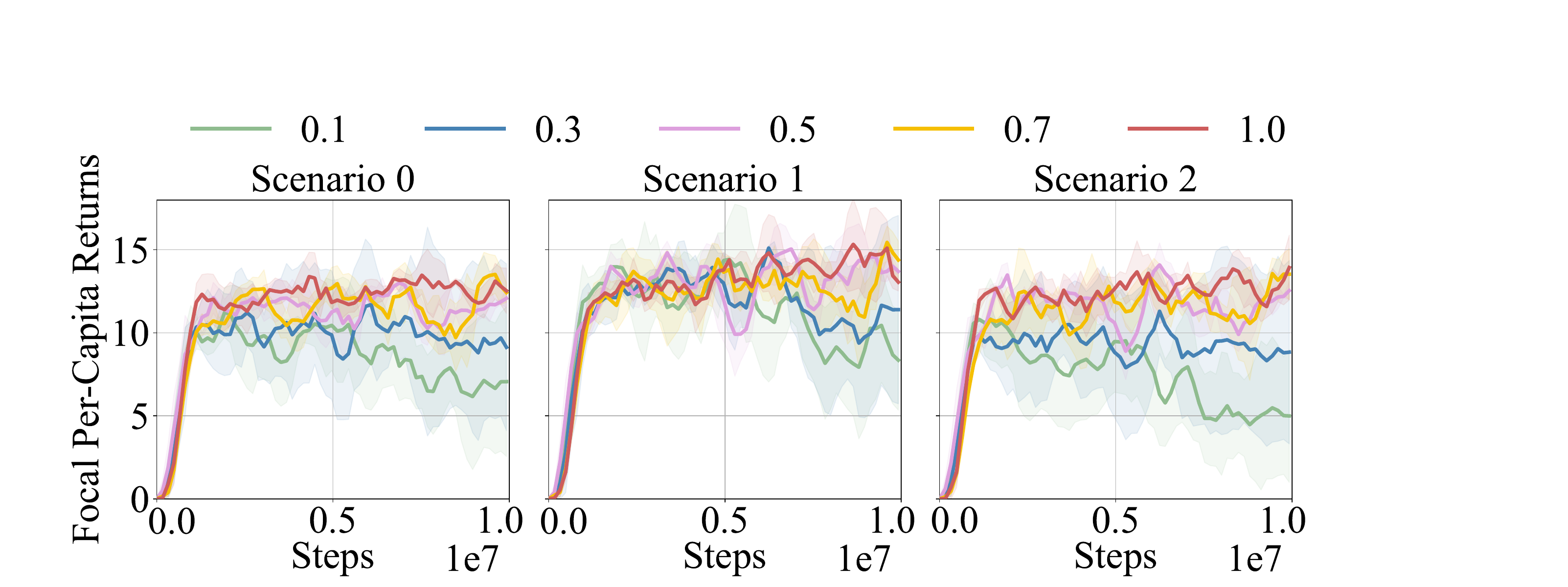} % Reduce the figure size so that it is slightly narrower than the column. Don't use precise values for figure width. This setup will avoid overfull boxes.
    \caption{Performance comparison of distilled policy networks with smaller network sizes on \emph{Pure Coordination} task.}
    \label{fig:net_size}
\end{figure}

To delve deeper into the resource efficiency of BiDist, we conducted experiments by progressively reducing the size of the distilled policy network. This reduction involved scaling down the width of the hidden layers within the distilled network to 0.7, 0.5, 0.3, and 0.1 times the size of the original learning policy network. The results are demonstrated in Figure~\ref{fig:net_size}
% One advantage of BiDist is that it requires maintaining only one distilled network, as opposed to the numerous rollout policy networks needed in the self-play approach. To delve into this aspect further, we conduct experiments by continuing to decrease the size of the distilled policy network.
% Specifically, we decrease the width of the hidden layers within the distilled network to 0.7, 0.5, 0.3, and 0.1 times the original learning policy network. The results are demonstrated in Figure~\ref{fig:net_size}. 

Notably, we observed that the performance of BiDist exhibits resilience when the size of the distilled policy network is maintained above 0.5. Thus, BiDist remains robust and effective even when employing a relatively smaller distilled policy network. However, a decline in the performance of BiDist becomes evident when the size of the distilled policy network falls below 0.5. This observation suggests that the distilled policy network needs to be of sufficient size to exhibit stronger representation ability, thereby enabling the retention of the longer-term data space. In our case, the boundary of the performance jump lies near the scale of 0.5. This insight implies that there is untapped potential for further optimizing resource utilization while maintaining performance. In summary, BiDist not only proves its resource efficiency through its single distilled network requirement but also demonstrates adaptability to smaller distilled network sizes.

\subsection{Distillation Intervals}

\begin{figure}[h]
    \centering
    \includegraphics[width=0.6\textwidth]{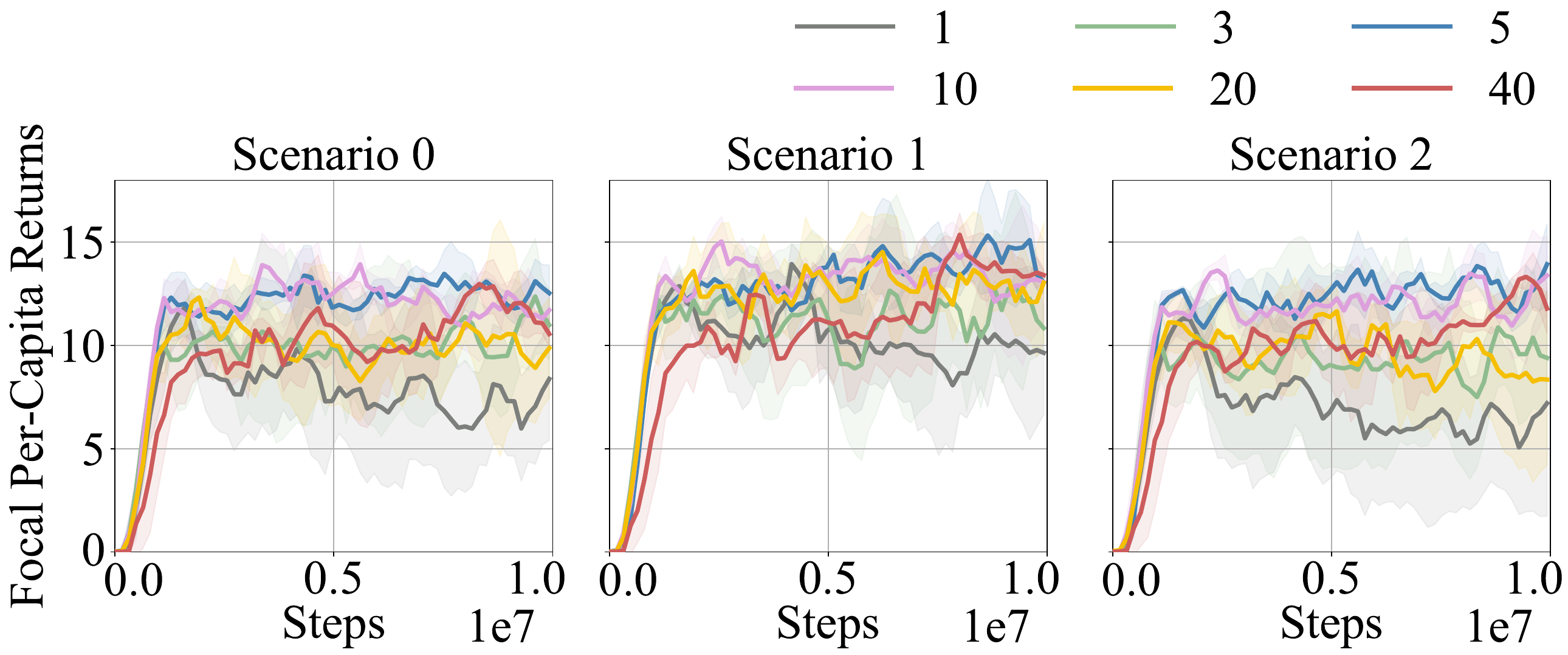} % Reduce the figure size so that it is slightly narrower than the column. Don't use precise values for figure width. This setup will avoid overfull boxes.
    \caption{Performance comparison of different distillation intervals on \emph{Pure Coordination} task. $k_d$ is set to 1, 3, 5, 10, 20, and 40, respectively.}
    \label{fig:interval}
\end{figure}

We empirically investigated the impact of different distillation intervals. Specifically, we set the distillation interval $k_d$ to 1, 3, 5, 10, 20, and 40, respectively. The corresponding results are illustrated in Figure~\ref{fig:interval}.

Overall, distillation intervals of $k_d=5$ and $k_d=10$ demonstrated relatively stable performance, with $k_d=5$ showing a slightly better outcome. Both smaller and larger distillation intervals led to varying degrees of performance degradation. It is observed that using a too-small distillation interval ($k_d=1$) overly frequently updates the distilled network, resulting in significant performance deterioration. Therefore, opting for a very small distillation interval is not advisable.  On the other hand, employing a large distillation interval ($k_d=40$) yields somewhat better results without encountering overfitting compared to $k_d=1$. However, the performance doesn't reach its optimum. Consequently, based on our empirical observations, we select $k_d=5$ as the distillation interval.
